%% file: main.tex
\documentclass[letterpaper, 10 pt, journal, twoside]{ieeetran}
\IEEEoverridecommandlockouts
\let\labelindent\relax

\input{math_commands.tex}

\usepackage{times}
\usepackage{amsmath,amssymb,amsfonts}
\usepackage{mathtools}
\usepackage{multicol}

\usepackage[shortlabels]{enumitem}
\usepackage{float}
\usepackage{graphicx}
\usepackage{xcolor}
\usepackage{siunitx}
\usepackage[font=footnotesize]{caption}
\usepackage{subcaption}
\usepackage[colorlinks,bookmarksopen,bookmarksnumbered,citecolor=black,urlcolor=black]{hyperref}
\usepackage[export]{adjustbox} 
\hypersetup
{
	pdftitle = {System Identification under Constraints and Disturbance: A Bayesian Estimation Approach},
	pdfauthor = {Sergi Martinez, Steve Tonneau, Carlos Mastalli},
	pdfauthor = {Author Names Omitted for Anonymous Review.},
	pdfsubject = {Submitted to IEEE Transation on Robotics (T-RO)},
	pdftoolbar = true,
	colorlinks = true,
	linkcolor = black,
	citecolor = black,
	urlcolor = black,
}

\usepackage{glossaries}
\usepackage{multirow}
\usepackage{hhline}
\usepackage{booktabs}
\usepackage{tikz}

\newcommand{\ra}[1]{\renewcommand{\arraystretch}{#1}}
\glsdisablehyper
\newacronym{sysid}{SysID}{system identification}
\newacronym{aba}{ABA}{articulated-body algorithm}
\newacronym{rnea}{RNEA}{recursive Newton--Euler algorithm}
\newacronym{oe}{OE}{optimal estimation}
\newacronym{com}{COM}{center of mass}
\newacronym{nlp}{NLP}{nonlinear programming}
\newacronym{kkt}{KKT}{Karush--Kuhn--Tucker}
\newacronym{ddp}{DDP}{differential dynamic programming}
\newacronym{mpc}{MPC}{model predictive control}
\newacronym{rl}{RL}{reinforcement learning}
\newacronym{imm}{IMM}{inertial matrix method}
\newacronym{id}{ID}{inverse dynamics}
\newacronym{fd}{FD}{forward dynamics}
\newacronym{ekf}{EKF}{extended Kalman filter}
\newacronym{lmi}{LMIs}{linear matrix inequalities}
\newacronym{map}{MAP}{maximum-a-posteriori}
\newacronym{qp}{QP}{quadratic program}

\usepackage{cleveref}
\Crefname{definition}{Definition}{Definitions}
\Crefname{proposition}{Proposition}{Propositions}
\Crefname{theorem}{Theorem}{Theorems}
\Crefname{figure}{Fig.}{Figs.}
\Crefname{equation}{Eq.}{Eqs.}
\Crefname{section}{Section}{Sections}
\Crefname{subsection}{Section}{Sections}
\Crefname{subsubsection}{Section}{Sections}
\Crefname{algorithm}{Algorithm}{Algorithms}

\newcommand{\orcid}[1]{\href{https://orcid.org/#1}{\includegraphics[width=0.6em]{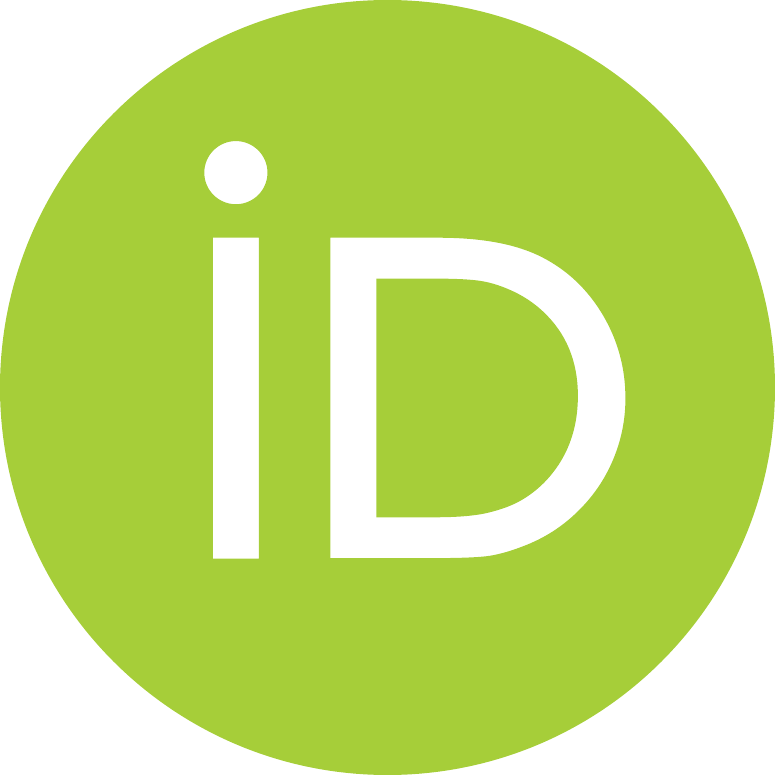}}}
\newcommand{\video}{https://youtu.be/ToTYp7mCkOQ}

\title{System Identification under Constraints and Disturbance: A Bayesian Estimation Approach}
\author{
    Sergi Martinez\orcid{}\quad
    Steve Tonneau\orcid{0000-0003-3001-8693}\quad
	Carlos Mastalli\orcid{0000-0002-0725-4279}
\thanks{This research was conducted as part of the Advancing MANipulation skills in Legged Robots (AMAN) project, a collaborative project supported by Tata Consultancy Services.
\textit{(Corresponding author: Carlos Mastalli)}}
\thanks{
Sergi Martinez and Carlos Mastalli are part of the Robot Motor Intelligence (RoMI) Lab, Heriot-Watt University, U.K.
Steve Tonneau is part of the Informatics School, University of Edinburgh, U.K.
}
}

\begin{document}

\maketitle

\begin{abstract} 
We introduce a Bayesian system identification (SysID) framework for jointly estimating robot’s state trajectories and physical parameters with high accuracy.
It embeds physically consistent inverse dynamics, contact and loop-closure constraints, and fully featured joint friction models as hard, stage-wise equality constraints.
It relies on energy-based regressors to enhance parameter observability, supports both equality and inequality priors on inertial and actuation parameters, enforces dynamically consistent disturbance projections, and augments proprioceptive measurements with energy observations to disambiguate nonlinear friction effects.
To ensure scalability, we derive a parameterized equality-constrained Riccati recursion that preserves the banded structure of the problem, achieving linear complexity in the time horizon, and develop computationally efficient derivatives.
Simulation studies on representative robotic systems, together with hardware experiments on a Unitree B1 equipped with a Z1 arm, demonstrate faster convergence, lower inertial and friction estimation errors, and improved contact consistency compared to forward-dynamics and decoupled identification baselines.
When deployed within model predictive control frameworks, the resulting models yield measurable improvements in tracking performance during locomotion over challenging environments.
\end{abstract}
\IEEEpeerreviewmaketitle
\input{chapters/intro}
\input{chapters/contributions}
\input{chapters/background}
\input{chapters/hybrid_dynamics}
\input{chapters/disturbed_dynamics}
\input{chapters/inertial_actuation}

\input{chapters/energy}
\input{chapters/PDDP}
\input{chapters/results}

\input{chapters/conclusions}
\appendices
\input{chapters/baumgarte_der}
\input{chapters/excitation}
\bibliographystyle{IEEEtran}
\bibliography{references}
\end{document}

%% file: math_commands.tex
\usepackage{amssymb}

\newcommand{\dynFunc}[1][]{\mathbf{f}_{#1}}

\newcommand{\eqFunc}[1][]{\mathbf{h}_{#1}}
\newcommand{\GeqFunc}[1][]{\mathbf{h}_{x_{#1}}}
\newcommand{\GineqFunc}[1][]{\mathbf{g}_{x_{#1}}}
\newcommand{\GeqFuncP}[1][]{\mathbf{h}_{p_{#1}}}
\newcommand{\GineqFuncP}[1][]{\mathbf{g}_{p_{#1}}}
\newcommand{\Integrator}[1][]{\boldsymbol{\Psi}_{#1}}
\newcommand{\obsFunc}[1][]{\mathbf{y}_{#1}}

\newcommand{\nx}{{n_{x}}} 
\newcommand{\nparams}{{n_{\theta}}} 

\newcommand{\nactparams}{{n_{\theta_a}}} 
\newcommand{\nq}{{n_{q}}} 
\newcommand{\nv}{{n_{v}}} 
\newcommand{\ntau}{{n_{u}}} 
\newcommand{\nz}{{n_{z}}} 
\newcommand{\nc}{{n_{c}}} 
\newcommand{\nb}{{n_{b}}} 

\newcommand{\arbitraryVar}{\mathbf{y}} 
\newcommand{\state}[1][]{\mathbf{x}_{#1}} 
\newcommand{\generalCtrl}[1][]{\boldsymbol{\tau}_{#1}} 
\newcommand{\inputVar}[1][]{\mathbf{u}_{#1}}
\newcommand{\ucertain}[1][]{\mathbf{w}_{#1}} 
\newcommand{\ucertainProj}[1][]{\mathbf{\tilde{w}}_{#1}} 
\newcommand{\params}[1][]{\boldsymbol{\theta}_{#1}} 
\newcommand{\dynparams}[1][]{\boldsymbol{\theta}_{{d}_{#1}}} 
\newcommand{\actparams}[1][]{\boldsymbol{\theta}_{{a}_{#1}}} 
\newcommand{\stateSeq}{\state[s]}

\newcommand{\inputSeq}{\inputVar[s]}

\newcommand{\spatialVel}[2][]{\prescript{#2}{}{\boldsymbol{v}_{#1}}}
\newcommand{\spatialAcc}[2][]{\prescript{#2}{}{\boldsymbol{a}_{#1}}}

\newcommand{\spatialVelLWA}[2][]{\prescript{#2}{}{\boldsymbol{\hat{v}}_{#1}}}
\newcommand{\pluckerTransform}[2][]{\prescript{#2}{}{\mathbf{X}_{#1}}}

\newcommand{\spatialFrame}[1][]{\mathcal{F}_{#1}}
\newcommand{\spatialPlacement}[2][]{\prescript{#2}{}{\mathbf{M}_{#1}}}
\newcommand{\lwaTransform}[1][]{\mathbf{\hat{M}}_{#1}}
\newcommand{\worldFrame}{\mathcal{W}}

\newcommand{\obsMeas}[1][]{\mathbf{\hat{y}}_{#1}} 
\newcommand{\posMeas}[1][]{\mathbf{\hat{q}}_{#1}} 
\newcommand{\velMeas}[1][]{\mathbf{\hat{v}}_{#1}} 
\newcommand{\accMeas}[1][]{\mathbf{\hat{a}}_{#1}} 
\newcommand{\angVelMeas}[1][]{\boldsymbol{\hat{\omega}}_{#1}} 
\newcommand{\ctrlMeas}[1][]{\boldsymbol{\hat{\tau}}_{#1}} 
\newcommand{\covariance}{\boldsymbol{\Sigma}}
\newcommand{\stateMean}[1][]{\mathbf{\bar{x}}_{#1}} 
\newcommand{\stateCov}[1][]{\covariance_{\mathbf{x}_{#1}}} 
\newcommand{\ucertainCov}[1][]{\covariance_{\ucertain[#1]}} 
\newcommand{\ucertainProjCov}[1][]{\covariance_{\ucertainProj[#1]}} 
\newcommand{\paramsMean}[1][]{\boldsymbol{\bar{\theta}}_{#1}} 
\newcommand{\paramsCov}[1][]{\covariance_{\boldsymbol{\theta}_{#1}}} 
\newcommand{\obsCov}[1][]{\covariance_{\obsMeas[#1]}} 

\newcommand{\pos}[1][]{\mathbf{q}_{#1}} 
\newcommand{\vel}[1][]{\mathbf{v}_{#1}} 
\newcommand{\postVel}{\mathbf{\dot{\pos}}^+} 
\newcommand{\preVel}{\mathbf{\dot{\pos}}^-} 
\newcommand{\acc}[1][]{\mathbf{\dot{v}}_{#1}} 
\newcommand{\eff}[1][]{\boldsymbol{\tau}_{#1}} 
\newcommand{\tauBias}{\eff[b]} 
\newcommand{\friction}{\eff[f]}
\newcommand{\wrch}[1][]{\boldsymbol{\lambda}_{#1}}
\newcommand{\imp}[1][]{\boldsymbol{\Lambda}_{#1}}
\newcommand{\impLWA}[1][]{\boldsymbol{\hat{\Lambda}}_{#1}}

\newcommand{\R}{\mathbb{R}} 
\newcommand{\confManif}{\mathcal{Q}} 
\newcommand{\confTManif}[1][\pos]{\mathcal{T}_{#1}\confManif} 
\newcommand{\stateManif}{\mathcal{X}} 
\newcommand{\stateTManif}[1][\state]{\mathcal{T}_{#1}\stateManif} 
\newcommand{\implConstManif}{\mathcal{C}} 
\newcommand{\implConstTManif}[1][\pos]{\mathcal{T}_{#1}\implConstManif} 

\newcommand{\obsManif}{\mathcal{Z}} 
\newcommand{\inertialManif}{\mathcal{I}}
\newcommand{\rotMatrix}[2][]{\prescript{#2}{}{\mathbf{R}_{#1}}}
\newcommand{\rotMatrixParam}{\boldsymbol{\omega}}

\newcommand{\implicitConst}[1]{\boldsymbol{\phi}_{\!#1}}
\newcommand{\implicitContact}{\implicitConst{c}}
\newcommand{\implicitLoop}{\implicitConst{k}}
\newcommand{\dimplicitConst}[1]{\mathbf{\dot{\boldsymbol{\phi}}}_{#1}}
\newcommand{\dimplicitContact}{\dimplicitConst{c}}
\newcommand{\dimplicitLoop}{\dimplicitConst{k}}
\newcommand{\ddimplicitConst}[1]{\mathbf{\ddot{\boldsymbol{\phi}}}_{#1}}
\newcommand{\ddimplicitContact}{\ddimplicitConst{c}}
\newcommand{\ddimplicitLoop}{\ddimplicitConst{k}}

\newcommand{\inertialParams}[1][]{\boldsymbol{\pi}_{#1}} 
\newcommand{\frictionParams}[1][]{\boldsymbol{\gamma}_{#1}} 
\newcommand{\massMatrix}{\mathbf{M}}
\newcommand{\gravTerm}{\mathbf{g}}
\newcommand{\coriolisGravTerm}{\mathbf{h}}
\newcommand{\motionJac}[2][]{\prescript{#2}{}{\mathbf{J}_{#1}}}
\newcommand{\dmotionJac}[2][]{\prescript{#2}{}{\mathbf{\dot{J}}_{#1}}}
\newcommand{\contactJac}{\motionJac[c]{}}
\newcommand{\loopJac}{\motionJac[k]{}}
\newcommand{\motionJacLWA}[1]{\mathbf{\hat{J}}_{#1}}
\newcommand{\contactJacLWA}{\motionJacLWA{c}}
\newcommand{\constraintForce}{\wrch{}}

\newcommand{\contactImpulse}{\imp[c]}
\newcommand{\biasAcc}[1][b]{\mathbf{a}_{#1}}
\newcommand{\biasAccLWA}[1][b]{\mathbf{\hat{a}}_{#1}}
\newcommand{\biasVel}[1][b]{\mathbf{v}_{#1}}

\newcommand{\mass}{m}
\newcommand{\lever}{\mathbf{h}}
\newcommand{\barycenter}{\mathbf{c}}
\newcommand{\rotInertia}{\mathbf{I}}
\newcommand{\rotInertiaBarycenter}{\rotInertia_c}
\newcommand{\eigInertia}{\mathbf{D}}
\newcommand{\regInertialMatrix}{\mathbf{Y}}
\newcommand{\regKineticMatrix}{\mathbf{Y}_K}
\newcommand{\regPotentialMatrix}{\mathbf{Y}_U}

\newcommand{\KineticEnergy}{K}
\newcommand{\PotentialEnergy}{U}
\newcommand{\MechanicalEnergy}{T}

\newcommand{\secMomentMass}{\mathbf{L}}
\newcommand{\residual}[1]{\mathbf{r}_{#1}}
\newcommand{\residualDyn}{\residual{d}}
\newcommand{\residualImpactDyn}{\residual{i}}
\newcommand{\residualAcc}{\residual{a}}
\newcommand{\residualVel}{\residual{v}}
\newcommand{\residualVelContact}{\residualVel^c}
\newcommand{\residualVelLoop}{\residualVel^k}
\newcommand{\residualAccContact}{\residualAcc^c}
\newcommand{\residualAccLoop}{\residualAcc^k}
\newcommand{\residualLWA}[1]{\mathbf{\hat{r}}_{#1}}

\newcommand{\residualVelLWA}{\residualLWA{v}}
\newcommand{\residualVelContactLWA}{\residualVelLWA^c}

\newcommand{\pvar}[1][]{\mathbf{p}_{#1}}

\newcommand{\valFunc}[1][]{\mathcal{V}_{#1}}
\newcommand{\qualFunc}[1][]{\mathbf{Q}_{#1}}
\newcommand{\qualFuncConst}[1][]{\Tilde{\mathbf{Q}}_{#1}}
\newcommand{\costGrad}[1][]{\boldsymbol{\ell}_{#1}}
\newcommand{\lagHess}[1][]{\mathcal{L}_{#1}}
\newcommand{\mulpDyn}[1][]{\boldsymbol{\gamma}_{#1}}
\newcommand{\mulpDynNext}[1][]{\boldsymbol{\gamma}_{#1}^{\mathbf{+}}}
\newcommand{\mulpEq}[1][]{\boldsymbol{\xi}_{#1}}
\newcommand{\mulpEqNext}[1][]{\boldsymbol{\xi}_{#1}^{\mathbf{+}}}
\newcommand{\ffPolicy}[1][]{\mathbf{k}_{#1}}

\newcommand{\fbPolicy}[1][]{\mathbf{K}_{#1}}
\newcommand{\pPolicy}[1][]{\mathbf{P}_{#1}}
\newcommand{\ffPolicyFinal}[1][]{\boldsymbol{\pi}_{#1}}
\newcommand{\fbPolicyFinal}[1][]{\boldsymbol{\Pi}_{#1}}
\DeclareDocumentCommand{\pPolicyFinal}{ O{} O{} }{\boldsymbol{\Pi}_{{p#1}_{#2}}}
\newcommand{\fbPolicyParam}{\boldsymbol{\Pi}_{\delta\state[0]}}

\newcommand{\dynFeas}[1][]{\mathbf{\bar{f}}_{#1}}

\newcommand{\eqFeas}[1][]{\mathbf{\bar{h}}_{#1}}
\newcommand{\ineqFeasP}[1][]{\mathbf{\bar{g}}_{p_{#1}}}
\newcommand{\eqFeasP}[1][]{\mathbf{\bar{h}}_{p_{#1}}}

\newcommand{\rankSpace}{\mathbf{Y}}
\newcommand{\rank}{\mathbf{y}}
\newcommand{\NullSpace}{\mathbf{Z}}
\newcommand{\Null}{\mathbf{z}}

\newcommand{\transpose}{\intercal}
\newcommand{\zeroVec}[1][]{\mathbf{0}_{#1}}
\newcommand{\eyeMatrix}[1][]{\mathbf{1}_{#1}}

\newcommand{\SO}[1][]{SO(#1)}
\newcommand{\SE}[1][]{SE(#1)}

\newcommand{\frobnorm}[1]{\left\lVert #1 \right\rVert_F}

%% file: chapters/intro.tex
\section{Introduction}
\begin{figure}[t]\centering
\href{\video&t=0}{\includegraphics[width=1.0\linewidth]{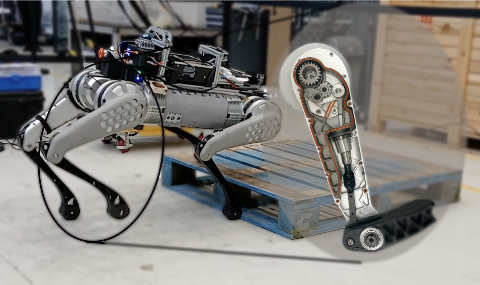}}
    \caption{
        The Unitree B1 quadruped robot performing a step-up maneuver. 
        The leg includes a reduction gear and a four-bar linkage, both explicitly modeled as constraints within our~\gls{sysid} framework. 
        The highlighted cross-section reveals the internal linkage bars, whose inertial properties and actuator-side friction effects are jointly identified to improve physical consistency and model fidelity.To watch the video, click the picture or see \texttt{\url{\video}}.
    }
    \vspace{-1.0em}
    \label{fig:cover}
\end{figure}

\IEEEPARstart{S}{ystem} identification~(\acrshort{sysid}) is essential for narrowing the \emph{Sim2Real} gap, enabling dynamic, robust locomotion and manipulation in robotics~\cite{bjelonic2025}.
Conventional \acrshort{sysid} methods typically assume a perfect knowledge of the robot's localization and adopt a purely frequentist formulation (e.g.,~\cite{gautier1992exciting,swevers1996expid,leboutet2021survey}).
However, these assumptions often limit identification accuracy---particularly when data are scarce, models are nonlinear, or uncertainty quantification is required (\Cref{fig:cover}).

In floating-base systems, localization and identification are inherently coupled: errors in model parameters (e.g., inertial mismatch, friction mismodeling) directly degrade state estimation,
as many localization methods---such as Kalman filtering or factor-graph optimization---rely on model-based contact detection.
Conversely, localization inaccuracies compromise model identification, since accurate estimates of positions, velocities, and accelerations are required to construct the joint-torque regressors used to estimate physical parameters.
As a result, decoupled approaches often fail to achieve the precision and robustness required for synthesizing agile locomotion in optimization or learning-based controllers such as~\gls{mpc} and~\gls{rl}.

Frequentist approaches estimate the ``true'' parameters by minimizing an objective function such as the least-squares prediction error or the likelihood, yielding a single best-fit parameter vector that maximizes agreement with the observed data.
However, when datasets are limited (e.g., short trajectories, insufficient excitation, or missing sensor data) or the model is imperfect (e.g., simplified friction models, unmodeled compliance, or sensor delays), these methods often overfit, resulting in poor generalization and unreliable identification.

In contrast, Bayesian approaches mitigate these issues by incorporating prior knowledge and inferring a distribution over the model parameters rather than a single point estimate.
This probabilistic formulation provides a principled framework for uncertainty quantification and prior integration.
Although Bayesian methods are typically more computationally demanding, their complexity can be substantially reduced by exploiting the Markovian and parametric structure of the dynamics—drawing inspiration from efficient solvers developed for~\gls{mpc} with full-system dynamics (e.g.,~\cite{koenemann-iros15,neunert-ral18,mastalli-invdynmpc}).

To address the aforementioned challenges, we formulate system identification within a Bayesian optimization framework relying on \emph{hybrid inverse dynamics} as follows:
\begin{subequations}\label{eq:oe_problem}
\begin{align}
\min_{\stateSeq, \inputSeq,\params}
& \frac{1}{2}\|\state[0]\ominus\stateMean[0]\|^2_{\stateCov[0]^{-1}}
 + \frac{1}{2}\|\params-\paramsMean\|^2_{\paramsCov^{-1}} \nonumber\\
& +\frac{1}{2}\sum_{k=0}^{N-1}\|\ucertain[k]\|^2_{\ucertainCov[N]^{-1}}
 +\frac{1}{2}\sum_{j=1}^{N}
 \|\obsMeas[j]\ominus\obsFunc(\state[j];\params\vert\ctrlMeas[j])\|^2_{\obsCov[j]^{-1}}
\end{align}
\begin{align}
\text{subject to}\quad
& \state[k+1]=\dynFunc(\state[k], \inputVar[k];\params\vert\ctrlMeas[k]),
  \label{eq:oe_problem:GEQ}\\
& \GeqFunc(\state[k],\inputVar[k];\params\vert\ctrlMeas[k])=\zeroVec,\\
& \GineqFunc(\state[k],\inputVar[k];\params\vert\ctrlMeas[k])\ge\zeroVec, \label{eq:oe_problem:GINEQ}\\
& \GeqFuncP(\params)=\zeroVec, \label{eq:oe_problem:PEQ}\\
& \GineqFuncP(\params)\ge\zeroVec. \label{eq:oe_problem:PINEQ}
\end{align}
\end{subequations}
where $\stateSeq = \{\state[0], \ldots, \state[N]\}$ denotes the state trajectory, $\inputSeq = \{\inputVar[0], \ldots, \inputVar[N-1]\}$ the input sequence, and $\params \in \R^{\nparams}$ the system parameters.
The state is defined as $\state=(\pos,\vel)\in\stateManif\subseteq\mathbb{R}^\nx$, where $\state[0]$ represents the arrival state, $\ucertain\in\stateTManif\subseteq\R^\nx$ denotes the process noise, and $\ctrlMeas\in\R^\ntau$ represents the measured control inputs.
During continuous-time phases (i.e., modes), $\inputVar$ aggregates the process noise, generalized accelerations, and constraint forces arising from contacts or closed-loop constraints, i.e.,
$\inputVar = (\ucertain, \acc, \constraintForce)$. 
For reset maps, $\inputVar$ instead comprises impulses, namely $\inputVar = \imp{}$.
The prior mean and covariance of the initial state, parameters and disturbaces are denoted by $(\stateMean[0],\stateCov[0])$, $(\paramsMean,\paramsCov)$, and $(\zeroVec,\ucertainCov)$, respectively.
The observation model $\obsFunc:\stateManif\times\R^{\nparams}\to\obsManif\subseteq\R^\nz$ maps the system state and parameters to the measurement space, with corresponding measurements $\obsMeas\in\obsManif$ and covariance $\obsCov$.
The equality and inequality constraints~\Cref{eq:oe_problem:GEQ}-(\ref{eq:oe_problem:PINEQ}) enforce dynamic consistency, physical feasibility, and parameter validity.

To evaluate the likelihood and enforce dynamic consistency in the optimization problem~\Cref{eq:oe_problem}, we rely on a physically consistent model of the robot’s dynamics, including kinematic loops, bilateral contacts, and discrete transitions (reset maps).
This model can be formulated using either forward or inverse dynamics.

Below, we identify the critical limitations of state-of-the-art~\gls{sysid} frameworks and explain how our approach overcomes them. 

\subsection{Forward and Inverse Dynamics}
Forward dynamics formulations are sensitive to coarse integration timesteps and can suffer from numerical instability, as frequently reported in the~\gls{mpc} literature~\cite{ereztrajectory,mastalli-invdynmpc}.
By decoupling the integrator from the dynamics, inverse dynamics formulations yield better-conditioned optimization problems, leading to faster convergence, improved robustness.
Inverse dynamics makes the physical-consistency relations explicit through equality constraints that couple motion, actuation inputs, and contact interactions.
Incorporating inverse dynamics within~\gls{sysid} frameworks can therefore enhance identification accuracy and physical consistency.
However, this formulation introduces additional decision variables compared to forward dynamics, increasing computational complexity despite its sparser structure.
Specifically, it augments the optimization problem with generalized accelerations and contact forces at each timestep, adding $N(\nv + \nc)$ decision variables.
To address this, we exploit the problem’s structured, allowing us to derive an efficient~\gls{sysid} optimization solver~(\Cref{sec:parametrized_riccati}).

\subsection{Kinematic Loops and Bilateral Constraints}
Beyond the inverse dynamics, the motion of a robot is also governed by constraints arising from intermittent contacts, closed kinematic loops, and discrete events such as touchdown and liftoff.
These phenomena restrict the admissible state-control space and introduce discontinuities in the dynamics.
When such effects are neglected, \gls{sysid} algorithms often compensate by distorting inertial or friction parameters, resulting in biased models and inaccurate state estimation.
Accurately enforcing contact constraints---especially during multi-contact motions---is therefore essential to preserve physical consistency and ensure reliable~\gls{sysid}.
Similarly, kinematic loops must be explicitly modeled, as their coupling structure allows friction to be correctly identified on the motor side rather than being absorbed at the joint level.
This improves parameter consistency across the transmission, but requires to develop analytical derivatives for mode and reset maps (\Cref{sec:derivatives_of_hybrid_dynamics_with_closed_loops}) and an especial treatment of external disturbance (\Cref{sec:disturbance_under_implicit_const}).

\subsection{Incorporating Physical and Parametric Constraints in Estimation}
Incorporating additional constraints provides a principled means to integrate prior knowledge, enforce physical consistency and hardware limitations within the~\gls{sysid} process. 
Such constraints serves a dual purpose: they encode prior information about the robot’s physical parameters while ensuring that the identified models remains dynamically and physical consistent.
For example, many robots exhibit discrete symmetries (such as left-right symmetry in bipeds), which can be imposed as equality constraints on the parameters~\cite{ordonez2024morphological}. 
Similarly, known quantities such as the total mass or bounds on friction coefficients can be enforced through equality or inequality constraints, thereby improving parameter identifiability and estimation accuracy (\Cref{sec:dyn_inertia}).
These constraints are formally represented by~\Cref{eq:oe_problem:PEQ} and (\ref{eq:oe_problem:PINEQ}), whose structured form can be exploited to improve computational efficiency particularly when running~\gls{sysid} over long or high-frequency datasets.
To fully exploit this structure, our framework combines Lie-algebra–based parameterizations (\Cref{sec:dyn_inertia}) with the advanced~\gls{sysid} optimization solver~(\Cref{sec:parametrized_riccati}).

\subsection{Friction and its Role in Energy Dissipation}
Traditional identification methods often neglects the joint friction, especially nonlinear effects such as stiction.
Friction introduces nonlinear, discontinuous, and hysteresis effects that are difficult to model accurately \cite{olsson1998friction}. 
This challenge is exacerbated in modern torque-controlled robots, which typically lack direct torque sensing---making only the combined effects of friction and limb inertia externally observable. 
Consequently, accurate identifying friction parameters becomes difficult, and localization must be tightly integrated within the overall~\gls{sysid} process.
To mitigate this issue, we leverage energy-based observations~(\Cref{sec:energy}), inspired by classical energy regressors \cite{gautier1997dynamic}. 
By explicitly incorporating measurements of input power and dissipated energy, our framework provides additional information to dissentagle friction effects from inertial dynamics, improving the identifiability of friction parameters even in the absence of direct torque measurements.

%% file: chapters/contributions.tex
\subsection{Contribution}
\begin{figure*}[t]\centering
    \href{\video&t=015}{\includegraphics[width = 1. \linewidth]{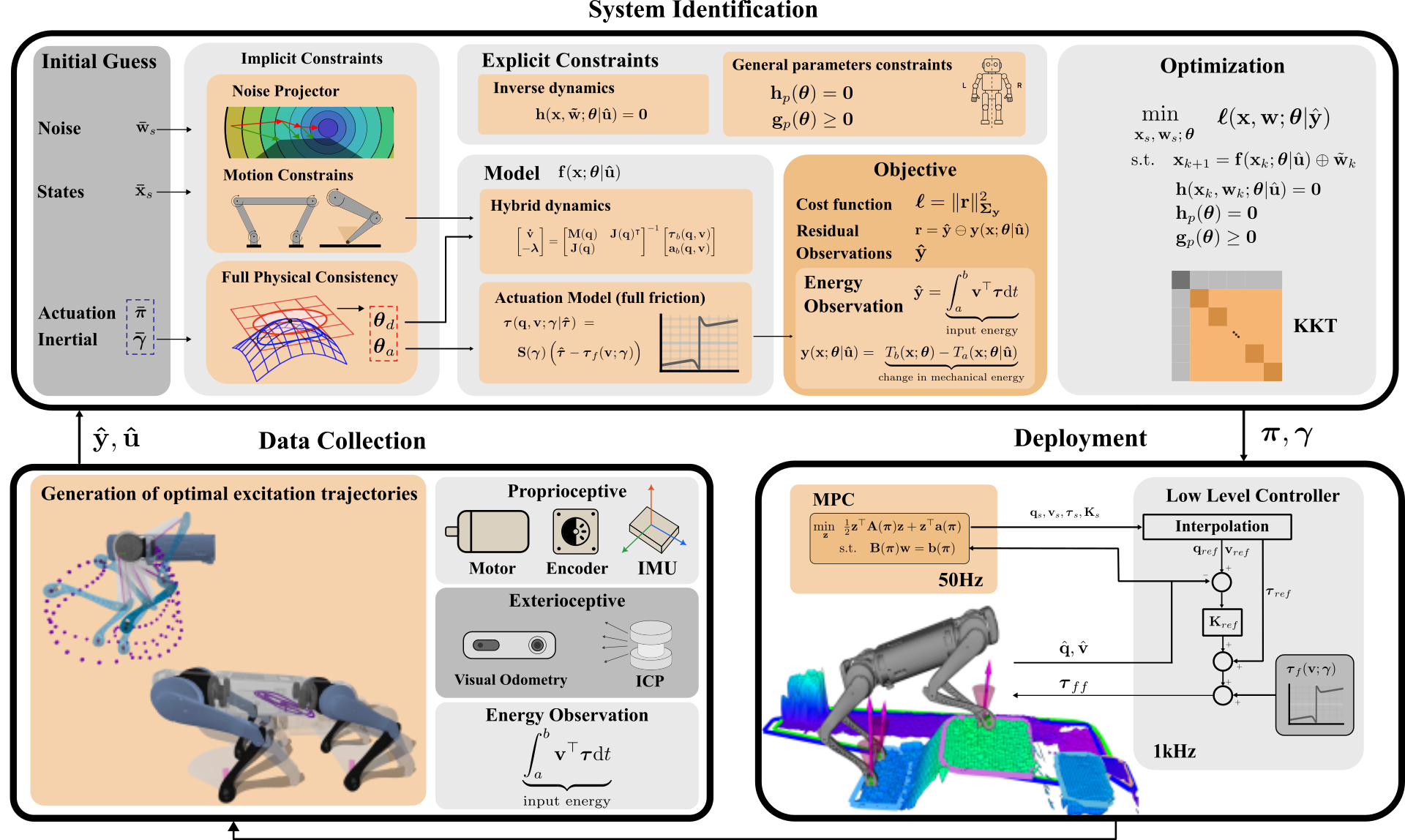}}
\caption{Overview of the our Bayesian optimization pipeline for system identification.
Proprioceptive measurements (e.g., encoders, IMU) and optional exteroceptive measurements (e.g., visual odometry/ICP) are synchronized and used to jointly estimate the state trajectory, disturbances, actuation effects (including friction), and physical parameters.
Inertial parameters are represented with a physically consistent parametrization, and process noise is projected onto the constraint-consistent tangent space to respect implicit motion constraints.
Dynamic consistency is enforced through explicit inverse-dynamics and motion constraints (contacts, closed-loop kinematics, hybrid/reset events), together with general parameter equality/inequality constraints.
Energy-based observations provide additional identifiability of friction by enforcing consistency between actuation power, dissipated energy, and changes in mechanical energy.
The resulting structured KKT system is solved efficiently with an equality-constrained Riccati approach, enabling scalable~\gls{sysid} over long, multi-rate datasets and deployment in control pipelines (e.g., MPC).To watch the video, click the picture or see \texttt{\url{\video}}.}
\label{fig:sysid_framework}
\end{figure*}

In this work, we introduce a unified and physically consistent framework for Bayesian system identification of floating-base robots subject to contact and loop constraints (\Cref{fig:sysid_framework}).
Our approach integrates accurate dynamic modeling, constraint-aware noise handling, and structure-exploiting optimization to achieve robust and energy-consistent identification across complex motion regimes.
Specifically, our framework:
\begin{enumerate}[(i)]
\item embeds a parameterized forward or inverse dynamics model with analytical derivatives,
\item handles closed-loop kinematics, bilateral contacts, and discrete reset maps in a unified formulation,
\item enforces dynamically consistent additive noise that respects contact and closed-loop constraints,
\item incorporates a smooth and differentiable friction model capturing stiction, dry, viscous, and Stribeck effects,
\item leverages energy-based observations to improve the accuracy and observability of identified dynamics, and
\item employs an efficient stagewise optimizer combined with manifold-based parameterizations to enforce physical consistency while maintaining computational scalability.
\end{enumerate}

Our work lies at the intersection of physically consistent identification~\cite{sousa2014lmi,wensing2018lmi,traversaro2016manifold}, constraint-aware estimation and identification under contact and loops~\cite{kolev2015physicallyconsistent,zhang2024constrainedsysid,khorshidi2024physconcontact,chignoli2023propagation}, and constraint-explicit optimal control for closed-chain mechanisms~\cite{dematteis2025parallel}.
We validate our~\gls{sysid} framework in simulated and real experiments across multiple robotics platforms, demonstrating consistent improvements in parameter accuracy over classical frequentist identification methods.

The following section reviews related research efforts in system identification and physically consistent estimation.

%% file: chapters/background.tex
\section{Related Work}\label{sec:related}
Reliable system identification in robotics builds upon decades of work spanning classical rigid-body regression, probabilistic estimation, and structure-exploiting optimization.
Yet, many state-of-the-art methods remain simplified; for example, they decouple localization from \gls{sysid}, rely predominantly on frequentist formulations, and often neglect energy consistency, friction, actuation effects, and closed-loop mechanisms.
This section reviews key developments relevant to our approach---covering traditional identification methods, joint estimation of states and parameters, probabilistic factor-graph frameworks, and recent advances in dynamics-constrained optimization and physically consistent modeling.

\subsection{Classical Identification Methods}
Early rigid body parameter identification writes inverse dynamics as a regression in the inertial parameters and solves least-squares from sufficiently excited trajectories~\cite{gautier1992exciting,swevers1996expid,mistry-inertiaest09,leboutet2021survey}. 
A key challenge is \emph{identifiability}: many parameters are linearly dependent, producing rank-deficient regressors and nonphysical solutions. 
Two main approaches address this issue. 
First, convex formulations using~\gls{lmi} enforce \emph{physical consistency} by constraining the identified parameters to correspond to realizable mass distributions~\cite{sousa2014lmi,wensing2018lmi}. 
Second, manifold parameterizations guarantees full physical consistency by construction, e.g., by encoding triangle inequalities on principal moments.
This allows smooth unconstrained optimization on minimal coordinates~\cite{martinez2025multi,traversaro2016manifold}. 
To improve robustness to feedback noise, \emph{closed-loop output-error} (CLOE/DIDIM) methods minimize simulation mismatch between measured and simulated torques in a closed-loop setting~\cite{gautier2010cloe}.
Nevertheless, these approaches assume accurate state and contact trajectories; even small pose or velocities biases can corrupt parameter identification, especially in contact-rich scenarios.
In this context, a promising alternative to traditional~\gls{sysid} approaches is Bayesian optimization.

\subsection{Joint State-Parameter Estimation}
To overcome state-parameter coupling, recent work performs joint estimation of localization, contact forces, and dynamics parameters within a unified optimization framework.
Kolev and Todorov~\cite{kolev2015physicallyconsistent} showed that, in contact-rich tasks, decoupled pipelines are fragile---millimeter-level state biases can make~\gls{sysid} ill-posed---while jointly optimizing trajectories, contact forces, and parameters yields consistent solutions.
Other examples include constrained least squares for identifying friction and motor inertia on the Digit humanoid~\cite{zhang2024constrainedsysid} and physically consistent contact identification on the Spot quadruped~\cite{khorshidi2024physconcontact}.
This latter example eliminates unknown contact forces through nullspace projections of whole-body dynamics while enforcing~\gls{lmi}-based physical consistency. 
To overcome these pitfalls,~\gls{sysid} frameworks that do not rely on known state trajectories are preferable alternatives.

\subsection{Probabilistic and Factor-Graph Formulations}
A common batch realization of joint estimation is~\gls{map} inference used on factor graphs.
Factor graphs fuse measurements with dynamics and kinematic models while maintaining handling manifold consistency via retractions~\cite{dellaert2017factor}.
Such frameworks naturally incorporate priors over both states and parameters, and can encode contact or loop-closure constraints directly as factors---reducing estimator bias compared with measurement-driven approaches.
However, generic factor-graph solvers often fail to exploit the sparsity and Markovian structure of rigid-body dynamics, resulting in slower convergence and the need for kinematic models only~\cite{dellaert2017factor}.
Scalability is particularly important in~\gls{sysid}, as joint friction identification often requires handling large, high-frequency datasets.

\subsection{Structure-Exploiting Solvers}
To achieve scalability, recent works draw inspiration from optimal control.
These methods exploit the Markovian structure of dynamics through Riccati recursion derived from the Bellman equation.
Algorithms such as DDP/iLQR handles constraints via projection or augmented Lagrangian penalties while retaining efficient fast Riccati recursion in their backward pass~\cite{tassa2014clddp,mastalli22auro,howell2019altro,jordana2023stagewise}. 
Related equality-constrained Riccati recursions further tighten the link between second-order methods and structure-exploiting linear algebra for constrained optimal control~\cite{vanroye2023ecriccati,mastalli-invdynmpc,parilli25-endpointmpc}. 
Although originally developed for control, these solvers are directly relevant to system identification: they preserve computational efficiency---which is important to handle large dataset---while enforcing physical and contact constraints within the estimation process.
However, efficient solvers for~\gls{sysid} require extending these by introducing a parametrized stagewise approach derived from the Bellman equation.

\subsection{Closed Kinematic Loops and Structural Constraints}
Many robotic platforms include closed kinematic loops and transmissions that violate open-chain assumptions~\cite{featherstone2014rigid}. 
Early work introduce elimination procedures to account for such loops~\cite{khalil1995closedloopbase,bennis1992icra,gautier1991jrs,shome1998general}, while more recent \emph{constraint embedding} formulations restore recursive efficiency by incorporating loop closure directly in the dynamics~\cite{chignoli2023propagation}.
This propagation-based view generalizes the~\gls{aba} to looped subsystem, improving computational efficiency and numerical stability. 
In control, explicitly enforcing constraints improve solution quality and expand feasible motion sets~\cite{dematteis2025parallel}.
These insights suggest that~\gls{sysid} pipelines should likewise \emph{embed} loop-closure constraints rather than approximate them as serial chains.
By leveraging hybrid dynamics and their analytical derivatives to handle bilateral contact, closed-loop mechanism, and reset maps, we strength the capabilities of current~\gls{sysid} frameworks.

\subsection{Energy- and Power-Based Identification}
While embedding loop and transmissions constraints improves modeling fidelity, observability can remain limited---particularly for frictional effects.
Energy-based formulations exploit work--energy principle to identify inertial and friction parameters without explicitly differentiating accelerations~\cite{caccavale1994energy,gautier1997powermodel}.
Such formulations are inherently energy-aware and often yield better-conditioned estimates.
When combined with physical-consistency constraints~\cite{sousa2014lmi,traversaro2016manifold,wensing2018lmi}, they enable parameter identification consistent with observed energy flow, even without direct force/torque sensing.
This offers a valuable signal source when internal actuator sensing is limited.
However, incorporating energy-based observations into Bayesian optimization poses significant challenges for computing analytical derivatives.

Next, we describe the modeling of hybrid dynamics and closed-loop mechanisms, along with the computation of their analytical derivatives.

%% file: chapters/hybrid_dynamics.tex
\section{Hybrid Dynamics with Closed-Loop Mechanisms}\label{sec:background}
\begin{figure*}[t]\centering
    \includegraphics[width = 0.24\linewidth]{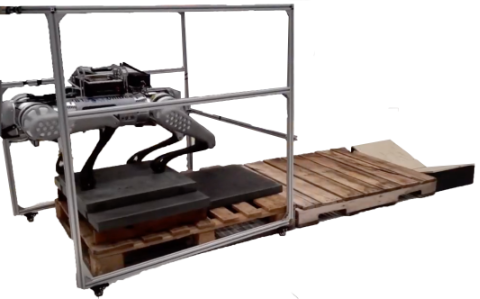}
    \includegraphics[width = 0.24\linewidth]{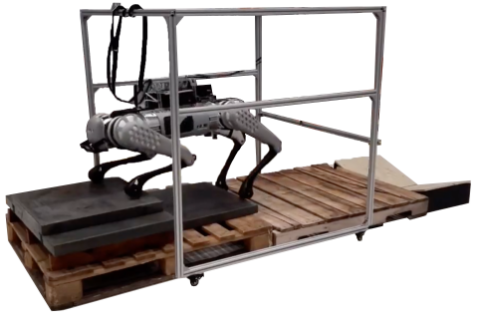}
    \includegraphics[width = 0.24\linewidth]{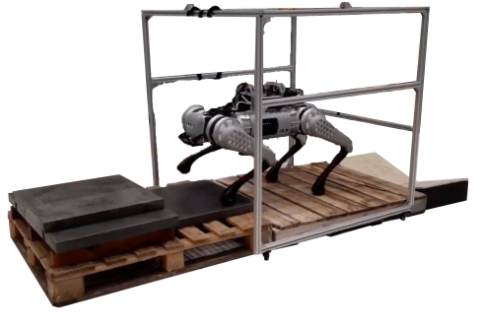}
    \includegraphics[width = 0.24\linewidth]{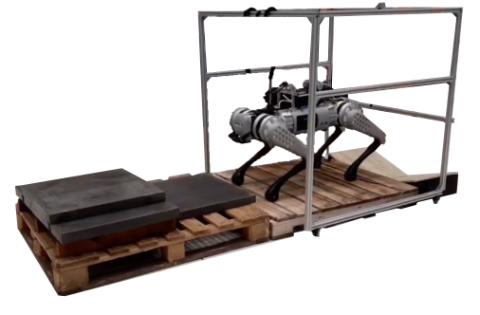}
    \href{\video&t=88}{\includegraphics[width = 0.19\linewidth]{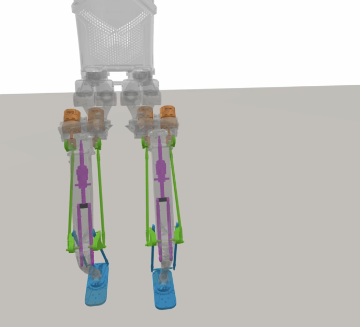}
    \includegraphics[width = 0.19\linewidth]{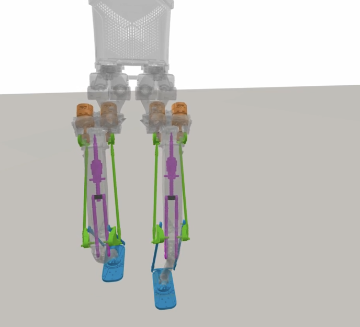}
    \includegraphics[width = 0.19\linewidth]{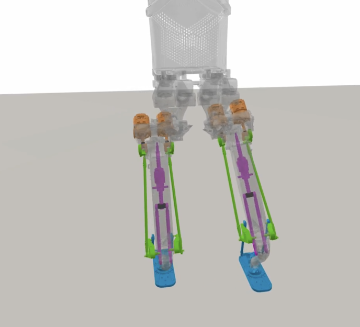}
    \includegraphics[width = 0.19\linewidth]{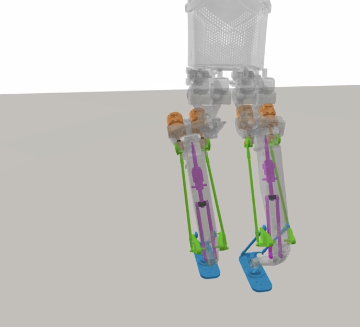}
    \includegraphics[width = 0.19\linewidth]{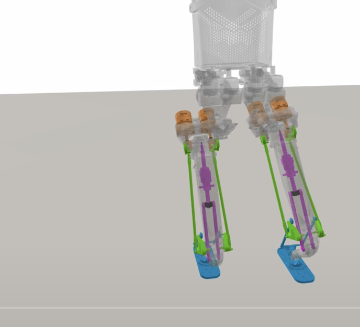}}
\caption{Top: Unitree B1 robot traversing a rough terrain. Bottom:  Kangaroo robot walking on a sidewalk with an identified model.
Planning, control, and estimation for these legged systems require accurately capturing contact dynamics and closed-loop mechanism effects inherent to multibody systems.}
\label{fig:snapshots}
\end{figure*}

Hybrid dynamics form the foundation of control, planning, and estimation in legged robotics (\Cref{fig:snapshots}). 
They capture the relationship between the forces acting on the system and the constraints they impose.

In this section, we introduce the mathematical formulation of hybrid dynamics, building on an introduction to rigid body algorithms and incorporating implicit constraints used to model bilateral contacts and closed-loop mechanisms.
We show how to efficiently compute the analytical derivatives of their modes and reset maps, which are key ingredients of our \gls{sysid} approach.

\subsection{Algorithms for Rigid Body Dynamics}
The dynamics of rigid-body systems are governed by the following differential equation:
\begin{align}\label{eq:dyn}
\massMatrix(\pos)\acc \;=\; \eff(\pos,\vel) - \coriolisGravTerm(\pos,\vel) + \motionJac[]{}(\pos)^\transpose\constraintForce,
\end{align}
where $\massMatrix:\confManif\to\R^{\nq\times\nq}$ is the joint-space inertia matrix; 
$\eff\in\R^\nq$ is the vector of generalized force modeling the actuation mechanism; 
$\coriolisGravTerm:\confManif\times\confTManif\to\R^\nq$ collects Coriolis, centrifugal, and gravity terms; 
and $\motionJac[]{}:\confManif\to\R^{\nc\times\nq}$ is the constraint Jacobian, mapping generalized accelerations to the constraint forces $\constraintForce\in\R^\nc$.
Moreover, $\pos\in\confManif\subseteq\R^{\nq}$ denotes the robot's configuration, $\vel\in\confTManif\subseteq\R^{\nq}$ the generalized velocity, 
and $\acc\in\stateTManif\subseteq\R^{\nq}$ the generalized acceleration, with $\nq$ defining the degrees of freedom of the system.

\begin{figure}[t]\centering
    \href{\video&t=100}{\begin{minipage}[t]{0.5\linewidth}
        \centering
        \includegraphics[width=\linewidth]{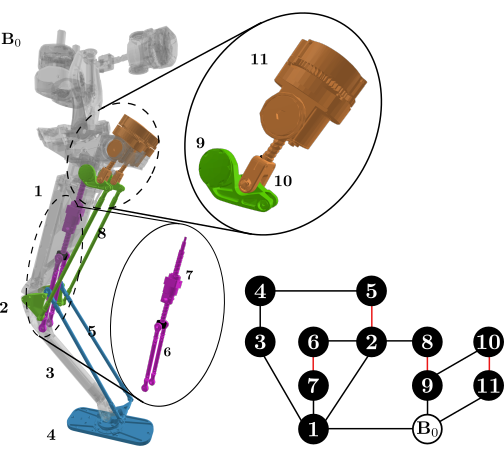}\\
        (a)
    \end{minipage}
    \begin{minipage}[t]{0.48\linewidth}
        \centering
        \includegraphics[width=\linewidth]{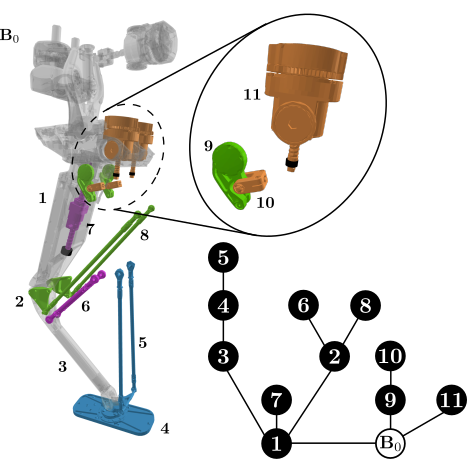}\\
        (b)
    \end{minipage}}
    \caption{Illustration of Kangaroo's leg kinematic structure and closed-loop modeling. 
(a) The actual mechanical design of the leg, including the actuator assembly, ankle linkage, and knee bars. 
The corresponding kinematic graph highlights the multiple closed-loop constraints present in the mechanism, resulting in a connectivity graph with cycles that prevent it from being a tree. 
(b) Representation of the same leg using an equivalent spanning tree, obtained by opening the closed loops. 
This process ensures that standard tree-based rigid-body algorithms can be applied.}
    \label{fig:kangaroo_mechanisms}
\end{figure}

When we evaluate~\Cref{eq:dyn}, in \emph{inverse dynamics} form, we compute $\eff$ given $(\pos,\vel,\acc)$ using the~\gls{rnea}.
For \emph{forward dynamics}, we compute $\acc$ from $(\pos,\vel,\eff)$ 
using the~\gls{aba}.
Both algorithms run in $\mathcal{O}(n)$ on kinematic trees because they exploit the branch-induced sparsity structure of the robot and fixed-size spatial operations~\cite{featherstone2014rigid,featherstone2000icra,luh1980rneA}.
The same structural properties also enable efficient computation of the analytical derivatives of~\gls{rnea} and~\gls{aba}~\cite{carpentier2018analytical}.
However, these algorithms apply only for robots whose connectivity graph forms a topological tree, i.e., a \emph{kinematic tree}.

In the following, we describe how to account for chords or cycles (see~\Cref{fig:kangaroo_mechanisms}), which arise when modeling bilateral contacts and closed-loop mechanisms, through implicit motion constraints.

\subsection{Dynamics with Implicit Motion Constraints}\label{sec:dyn_cons}
We refer to \emph{implicit motion constraints} as restrictions on the admissible motions that do not appear explicitly in~\Cref{eq:dyn} but arise from the system's structure or its interaction with the environment~\cite{featherstone2014rigid}.
Such constraints capture chords or cycles in the connectivity graph, which typically emerge when modeling bilateral contacts or closed-loop mechanisms.
In hybrid dynamics, the most relevant cases include:
(i) manifold-valued configurations (e.g., $\SO[3]$ for orientations), 
(ii) bilateral contact conditions,
(iii) holonomic constraints from structural closed-loop kinematics, and 
(iv) hybrid \emph{reset maps} at discrete events (e.g., touchdown, liftoff, impacts).
While we handle manifold-valued configurations using Lie algebra~\cite{sola2018micro}, the remaining constraints must be explicitly enforced in the dynamics.

Motion constraints of the form $\implicitConst{}(\pos)=\zeroVec$ describe holonomic and nonholonomic restrictions at the configuration level.
We can also express these constraints at velocity and acceleration levels:
\begin{subequations}\label{eq:kin_cons}
\begin{align}
\dimplicitConst{}(\pos,\vel) = \motionJac[]{}(\pos)\vel &= \zeroVec,  &\text{(velocity)} \\
\ddimplicitConst{}(\pos,\vel,\acc) = \motionJac[]{}(\pos)\acc &+ \mathbf{\dot{\motionJac[]{}}}(\pos,\vel)\vel = \zeroVec, &\text{(acceleration)}
\end{align}
\end{subequations}
where $\motionJac[]{}(\pos)=\partial \implicitConst{}/\partial \pos$ is the constraint Jacobian, and $\mathbf{\dot{\motionJac[]{}}}(\pos,\vel)$---often called the \emph{bias term}---captures its time variation along the motion.
This Jacobian can represent, for instance, bilateral contact constraints $\contactJac(\pos)=\partial \implicitContact/\partial \pos$ or closed-loop kinematics constraints $\loopJac(\pos)=\partial \implicitLoop/\partial \pos$.

These relations define the tangent space of feasible motions at $(\pos,\vel)$. 
In the presence of these equality constraints, the system's admissible velocities and accelerations must satisfy the algebraic conditions in~\Cref{eq:kin_cons}.
We formalize this within the hybrid dynamics framework by applying the Gauss's principle to both modes and reset maps, as described next.

\subsubsection{Motion constraints within modes}
To compute physically consistent accelerations and constraint forces, we apply Gauss's principle of least constraint~\cite{udwadia1992new}.
According to this principle, the actual acceleration is the one that minimizes the deviation from the unconstrained dynamics, subject to the set of admissible accelerations, i.e.,
\begin{equation}\label{eq:gauss_acc}
\begin{aligned}
\min_{\acc} \quad
    & \tfrac{1}{2}\,
      \bigl\|
        \acc - \acc^{\text{free}}(\pos,\vel,\eff(\pos,\vel))
      \bigr\|^2_{\massMatrix(\pos)} \\
\text{subject to} \quad
    & \motionJac{}(\pos)\,\acc = \biasAcc(\pos,\vel).
\end{aligned}
\end{equation}
where $\acc[\text{free}] \coloneqq \massMatrix(\pos)^{-1}\big(\eff(\pos,\vel) - \coriolisGravTerm(\pos,\vel)\big)$ denotes the unconstrained acceleration, $\biasAcc(\pos,\vel) = - \mathbf{\dot{\motionJac[]{}}}(\pos,\vel)\vel + \biasAcc^{*}(\pos,\vel)$ denotes the \emph{bias acceleration}, and $\biasAcc^{*}(\pos,\vel)$ specifies the desired constraint correction.
In practice, $\biasAcc^{*}(\pos,\vel)$ is often introduced through Baumgarte stabilization~\cite{baumgarter1972} to eliminate deviations from the target constraint motion.

From~\gls{kkt} conditions of~\Cref{eq:gauss_acc}, we obtain the following system of equations:
\begin{align}\label{eq:kkt_acc}
\begin{bmatrix}
\acc \\ -\constraintForce
\end{bmatrix}
=
\begin{bmatrix}
\massMatrix(\pos) & \motionJac[]{}(\pos)^\transpose \\
\motionJac[]{}(\pos) &
\end{bmatrix}^{-1}
\begin{bmatrix}
\tauBias(\pos,\vel) \\
\biasAcc(\pos,\vel)
\end{bmatrix}.
\end{align}
where $\tauBias(\pos,\vel)=\eff(\pos,\vel)-\coriolisGravTerm(\pos,\vel)$ as the \emph{force-bias term}, which includes actuation forces, joint friction, Coriolis and gravitational forces.
This expression can be derived by applying the Newton's method over the~\gls{kkt} residuals:
\begin{align}
\mathbf{r} =
\begin{bmatrix}
\residualDyn \\ \residualAcc
\end{bmatrix} =
\begin{bmatrix}
\massMatrix(\pos)\acc + \coriolisGravTerm(\pos,\vel) - \motionJac[]{}(\pos)^\transpose \constraintForce - \eff(\pos,\vel,\acc) \\
\motionJac[]{}(\pos)\acc - \biasAcc(\pos,\vel)
\end{bmatrix},
\end{align}
where $\residualDyn\in\R^\nq$ and $\residualAcc\in\R^\nc$ are the KKT residuals associated to the dynamics and acceleration-level constraints, respectively.
This formulation simultaneously computes the generalized accelerations $\acc$ and constraint forces $\constraintForce$, ensuring consistency with the holonomic and nonholonomic constraints defined by bilateral contacts, closed-loop mechanisms, etc.
The vector $\constraintForce\in\R^\nc$ corresponds to the Lagrange multipliers of the motion constraints, enforcing that the resulting accelerations remains on the constraint manifold $\implConstManif = \{\pos\in\confManif\subseteq\R^{\nq} \;|\; \implicitConst{}(\pos) = \zeroVec\in\R^{\nc}\}$.

\subsubsection{Motion constraints within reset maps}
During a hybrid event (impact, contact engage/disengage), the Newton--Euler equations are integrated over an infinitesimal time interval $[t^-,t^+]$.
By analogy with continuous modes, the Gauss impact principle defines the impact law as the post-impact velocity $\postVel$ that deviates as little as possible from pre-impact velocity $\preVel$:
\begin{equation}\label{eq:gauss_vel}
\begin{aligned}
\min_{\postVel} \quad
    & \tfrac{1}{2}\,
      \left\|
        \postVel - \preVel
      \right\|^2_{\massMatrix(\pos)} \\
\text{subject to} \quad
    & \motionJac[]{}(\pos)\,\postVel = \biasVel(\pos).
\end{aligned}
\end{equation}
where $\biasVel$ encodes the constraint velocity after impact (i.e., \emph{bias velocity}).
From the~\gls{kkt} conditions of~\Cref{eq:gauss_vel}, we obtain the system
\begin{align}
\label{eq:kkt_vel_reset}
\begin{bmatrix}
\postVel \\ -\imp{}
\end{bmatrix}
=
\begin{bmatrix}
\massMatrix(\pos) & \motionJac[]{}(\pos)^\transpose \\
\motionJac[]{}(\pos) &
\end{bmatrix}^{-1}
\begin{bmatrix}
\massMatrix(\pos)\preVel \\ \biasVel(\pos)
\end{bmatrix},
\end{align}
with $\imp{}$ is the generalized constraint impulse.
It arises from the following KKT residuals:
\begin{align}
\mathbf{r} =
\begin{bmatrix}
\residualImpactDyn \\ \residualVel
\end{bmatrix} =
\begin{bmatrix}
\massMatrix(\pos)(\postVel - \preVel) - \motionJac[]{}(\pos)^\transpose\imp{} \\
\motionJac[]{}(\pos)\postVel - \biasVel(\pos)
\end{bmatrix},
\end{align}
where $\residualImpactDyn\in\R^\nq$ and $\residualVel\in\R^\nc$ are the~\gls{kkt} residuals associated to impact law and velocity-level constraints, respectively.
This directly yields the classical impulse-momentum relation:
\begin{align}\label{eq:impulse_momentum_constraint}
\massMatrix(\pos)\big(\postVel-\preVel\big) \;=\; \motionJac[]{}(\pos)^\transpose \imp{}.
\end{align}
For bilateral contacts, the bias velocity recovers the standard \emph{impulse-momentum law}, i.e.,
\begin{align}
{\biasVel}_c(\pos) = -\epsilon\contactJac(\pos)\preVel,
\end{align}
with $\contactImpulse\geq\zeroVec$, $\epsilon\in[0, 1]$ the coefficient of restitution, where $\epsilon=0$ corresponds to a perfectly plastic impact and $\epsilon=1$ to a perfectly elastic impact. 

While the modeling of bilateral constraints is well established, computing analytical derivatives for systems with closed-loop kinematics remains nontrivial and is often a source of numerical and implementation complexity.

\section{Analytical Derivatives of Hybrid Dynamics with Closed-Loop Mechanism}\label{sec:derivatives_of_hybrid_dynamics_with_closed_loops}
Bayesian estimation in~\gls{sysid} require the sensitivities of the acceleration and constraint forces in~\Cref{eq:kkt_acc} (or velocity and impulse constraints) with respect to the generalized position $\pos$, velocity $\vel$, and dynamics parameters $\params$.
These sensitivities can be obtained by differentiating the KKT system in ~\Cref{eq:kkt_acc} in a forward-mode manner.
For an arbitrary variable $\arbitraryVar$, we have:
\begin{align}\label{eq:derivative_kkt_acc}
\begin{bmatrix}
\frac{\partial\acc}{\partial\arbitraryVar} \\ -\frac{\partial\constraintForce}{\partial\arbitraryVar}
\end{bmatrix} =
-\begin{bmatrix}
\massMatrix(\pos) & \motionJac[]{}(\pos)^\transpose \\
\motionJac[]{}(\pos) &
\end{bmatrix}^{-1}
\begin{bmatrix}
    \frac{\partial\residualDyn(\pos,\vel)}{\partial\arbitraryVar} \\
    \frac{\partial\residualAcc(\pos,\vel)}{\partial\arbitraryVar}
\end{bmatrix},
\end{align}
which is equivalent to applying the implicit function theorem as introduced in~\cite{mastalli-icra20}.
The right-hand-side terms can be computed analytically, i.e., $\partial\acc = -{\nabla\residual{}}^{-1}\partial\residual{}$ and $\partial\constraintForce = {\nabla\residual{}}^{-1}\partial\residual{}$, where $\partial\residual{}$ contains the partial derivatives of both dynamics and acceleration residuals ($\residualDyn$, $\residualAcc$).
In particular,
\begin{align}
    &\frac{\partial\residualDyn}{\partial\arbitraryVar} = \frac{\partial \mathrm{RNEA}(\pos,\vel,\acc)}{\partial\arbitraryVar} - \frac{\partial\eff(\pos,\vel,\acc)}{\partial\arbitraryVar},
\end{align}
where $\arbitraryVar \in {\pos,\vel,\params}$, and $\params\in\R^{\nparams}$ denotes the vector of dynamics parameters, including inertial, actuation, and joint-friction parameters.
The terms $\partial\mathrm{RNEA}/\partial\pos$ and $\partial\mathrm{RNEA}/\partial\vel$ correspond to the analytical derivatives of the~\gls{rnea}, as derived in~\cite{carpentier2018analytical}.

\subsubsection{Analytical derivatives within reset maps}
Analogously to~\Cref{eq:derivative_kkt_acc}, the forward-mode derivatives of the reset maps (i.e.,~\Cref{eq:kkt_vel_reset}) are computed as
\begin{align}\label{eq:derivative_kkt_vel}
\begin{bmatrix}
\frac{\partial\postVel}{\partial\arbitraryVar} \\ -\frac{\partial \imp{}}{\partial\arbitraryVar}
\end{bmatrix}
=
-\begin{bmatrix}
\massMatrix(\pos) & \motionJac[]{}(\pos)^\transpose \\
\motionJac[]{}(\pos) &
\end{bmatrix}^{-1}
\begin{bmatrix}
\frac{\partial \residualImpactDyn(\pos)}{\partial\arbitraryVar} \\
\frac{\partial \residualVel(\pos)}{\partial\arbitraryVar}
\end{bmatrix},
\end{align}
where the term $\frac{\partial\massMatrix(\pos)}{\partial\arbitraryVar}$ inside $\frac{\partial\residualImpactDyn}{\partial\arbitraryVar}$ is a third-order tensor.
To avoid explicitly forming such tensors, the upper block of the right-hand side vector in~\Cref{eq:derivative_kkt_vel} can be computed more efficently by exploting the identity
\begin{align}\label{eq:derivative_impulse}
\frac{\partial \residualImpactDyn}{\partial\arbitraryVar} = \frac{\partial\mathrm{RNEA}(\pos,\zeroVec,\postVel-\preVel)}{\partial\arbitraryVar} - \frac{\partial\gravTerm(\pos)}{\partial\arbitraryVar},
\end{align}
where, for $\arbitraryVar\in\pos,\vel,\params$, the analytical derivatives are computed as described in~\cite{carpentier2018analytical}.

In the following, we outline how to compute the analytical derivatives of the acceleration bias $\biasAcc$ (and the velocity bias $\biasVel$ for reset maps) in the presence of bilateral contacts and closed-loop mechanisms. 
For the actuation forces $\eff$ incorporating Coulomb, viscous, and Stribeck friction effects, we later describe how to compute the corresponding derivatives.

\subsection{Bilateral Contacts and Closed-Loop Mechanisms} \label{sec:contact_closed}
When the robots contains closed mechanisms, the algebraic loop-closure constraints $\implicitConst{k}(\pos)=\zeroVec$ can often be formulated in multiple ways.
This results in different spaning trees, different dimensions on the minimal coordinates set, and varying number of motion constraints.
For example,~\Cref{fig:4bar} illustrates this effect on a four-bar mechanism.

A similar situation arises in the case of bilateral contacts, which $\implicitContact(\pos)=\zeroVec$ is expressed as equality constraints defined by a reference frame placement.
In general, we define both bilateral contacts and loop-closure constraints as follows:
\begin{align}\label{eq:holonomic_constraints}\nonumber
\implicitContact(\pos) \;&=\; \mathrm{Log}(\spatialPlacement[1]{c}^*\ominus\spatialPlacement[1]{c}),  &\text{(bilateral contact)}\\
\implicitLoop(\pos) \;&=\; \mathrm{Log}(\spatialPlacement[{k_1}]{{k_2}}),  &\text{(loop closure)}
\end{align}
where $\spatialPlacement[1]{c}$, $\spatialPlacement[{k_1}]{{k_2}}$ are the placement of the contact constraint frame and the relative placement of the loop constraint frame respectively.
These placements are computed by the forward kinematics, and the superscript $*$ denotes the reference placement, and $\mathrm{Log}$ is the logarithm map from $\SE[3]$ to $\R^6$.

\subsubsection{Contacts and closed-loop mechanisms within reset maps}
Differentiating the holonomic constraints yields the velocity-level conditions for bilateral contacts and closed-loop mechanisms, which appears in the~\gls{kkt} residuals (i.e., $\residualVel=\dimplicitConst{}$) of reset maps:
\begin{align}\nonumber
\dimplicitContact(\pos,\vel) \;&=\; \pluckerTransform[1]{c}\spatialVel[1]{}^+ - \epsilon\pluckerTransform[1]{c}\spatialVel[1]{}^-, &\text{(bilateral contact)}\\\nonumber
\dimplicitLoop(\pos,\vel) \;&=\; \spatialVel[1]{k1}^+ - \pluckerTransform[k2]{k1}(\pos)\spatialVel[2]{k2}^+, &\text{(loop closure)}
\end{align}
with:\vspace{-0.8em}
\begin{align}\label{eq:dholonomic_constraints}
\spatialVel[1]{k1}^+ = \pluckerTransform[1]{k1}\spatialVel[1]{}^+, \quad \spatialVel[2]{k2}^+ = \pluckerTransform[2]{k2}\spatialVel[2]{}^+,
\end{align}
where $\pluckerTransform[1]{c}$, $\pluckerTransform[1]{k1}$, and $\pluckerTransform[2]{k2}$ are the Plücker transforms (i.e., the adjoint matrix of $\SE[3]$) mapping spatial velocities from local coordinates to $\spatialFrame[c]$, $\spatialFrame[k1]$ and $\spatialFrame[k2]$ coordinates, respectively; $\pluckerTransform[k2]{k1}(\pos)$ maps spatial velocities from $\spatialFrame[k2]$ to $\spatialFrame[k1]$ and depends on the robot configuration $\pos$.
The terms $\spatialVel[1]{}$ and $\spatialVel[2]{}$ denote the spatial velocities of the first and second bodies, expressed in local frames.
They can be computed after and before impact, i.e., $\spatialVel[]{}^+$ and $\spatialVel[]{}^-$, respectively.

Expressing the holonomic constraints in terms of the generalized velocities $\vel$ leads to the definition of the constraint Jacobian:
\begin{align}\nonumber
\contactJac(\pos) \;&=\; \pluckerTransform[1]{c}\,\motionJac[1]{}(\pos), &\text{(bilateral contact)}\\\nonumber
\loopJac(\pos) \;&=\; \motionJac[1]{k1}(\pos) - \pluckerTransform[k2]{k1}(\pos)\,\motionJac[2]{k2}(\pos),&\text{(loop closure)}
\end{align}
with:\vspace{-0.8em}
\begin{align}\label{eq:constraintJacobian}
\motionJac[1]{k1} = \pluckerTransform[1]{k1}\motionJac[1]{}, \quad \motionJac[1]{k2} = \pluckerTransform[2]{k2}\motionJac[2]{},
\end{align}
where $\motionJac[1]{}$ and $\motionJac[2]{}$ are the local Jacobians of the bodies $1$ and  $2$, respectively.

\begin{figure}[t]\centering
    \begin{minipage}[t]{0.48\linewidth}
        \centering
        \includegraphics[width=\linewidth]{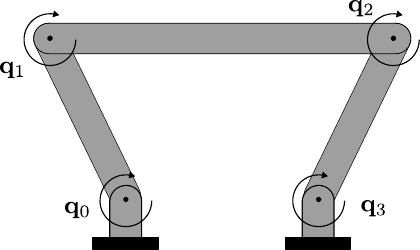}\\
        (a)
    \end{minipage}\hfill
    \begin{minipage}[t]{0.48\linewidth}
        \centering
        \includegraphics[width=\linewidth]{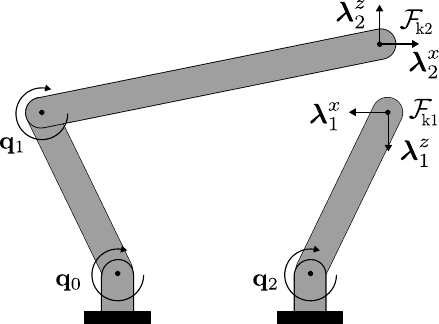}\\
        (b)
    \end{minipage}\hfill
    \vspace{0.5em}
    \begin{minipage}[t]{0.48\linewidth}
        \centering
        \includegraphics[width=\linewidth]{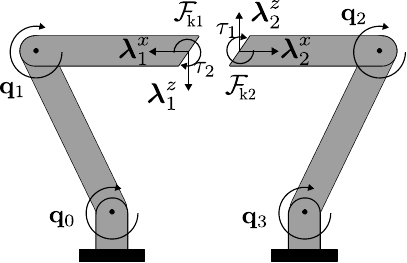}\\
        (c)
    \end{minipage}
    \begin{minipage}[t]{0.35\linewidth}
        \centering
        \includegraphics[width=\linewidth]{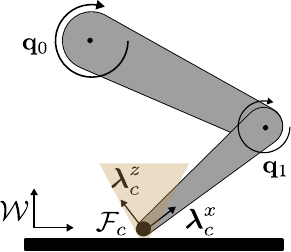}\\
        (d)
    \end{minipage}
    \caption{
        Illustration of implicit constraints: 
            (a) Standard four-bar mechanism, showing the loop closure; 
            (b) Mechanism opened at one joint, breaking the loop and reducing the system to a serial chain with two additional constraint forces required to maintain the original motion; 
            (c) Mechanism with one bar split in half, preserving the original degrees of freedom but now requiring two linear constraint forces and an additional torque to enforce the closed-chain kinematics; 
            (d) Planar leg in contact with the ground, the bilateral constraint is modeled exactly the same as in the closed loop mechanism but with the force affecting only one body. The contact frame $\spatialFrame[c]$ expressed in local coordinates does not align with the normal of the surface, world coordinates $\worldFrame$.
    }
    \label{fig:4bar}
\end{figure}

These Jacobians map $\vel\in\confTManif\subseteq\R^{\nq}$ to their spatial velocities expressed in their own body frames.
However, this is not convenient for modeling bilateral contacts (see~\Cref{fig:4bar}d).
This is because the friction cone must be enforced on contact impulses (i.e., constraints multipliers $\imp{}$) aligned with the world frame $\worldFrame$.
To this end, we define a configuration-dependent transformation that expresses the contact frame in world orientation:
\begin{align}\label{eq:lwaTransform}
\lwaTransform[1](\pos) = \big(\rotMatrix[1]{\worldFrame}{(\pos)},\, \zeroVec\big)\in\SE[3],
\end{align}
where $\rotMatrix[1]{\worldFrame}$ is the rotation from the local coordinates of body $1$ to the world frame $\worldFrame$, leading the world-aligned transformation $\lwaTransform[1]$.
Applying this transform yields the world-aligned Jacobian of  bilateral constraints:
\begin{align}\label{eq:lwaJac}
\contactJacLWA(\pos) \;&=\; \lwaTransform[1](\pos)\pluckerTransform[1]{c}\,\motionJac[1]{},
\end{align}
which allows us to compute the world-aligned impulses $\impLWA{}$ in~\Cref{eq:kkt_vel_reset}.
It also provides a direct way to express the world-aligned spatial velocities as $\spatialVelLWA[1]{}=\lwaTransform[1](\pos)\spatialVel[1]{}$, where the $\mathbf{\hat{\cdot}}$ notation to denote world-aligned quantities.

Velocity-based constraints depend only on the robot configuration.
Therefore, we only need to compute the derivatives of the velocity residuals with respect to position (i.e., $\partial\residualVel/\partial\pos$).
These derivatives, expressed in the local frame, are given by
\begin{align}\label{eq:velDerivatives}\nonumber
\frac{\partial\residualVelContact}{\partial\pos} \;&=\;\pluckerTransform[1]{c}\,\frac{\partial\spatialVel[1]{}^+}{\partial\pos}, &\text{(bilateral contact)}\\\nonumber
\frac{\partial\residualVelLoop}{\partial\pos} \;&=\; 
\frac{\partial\spatialVel[1]{k1}^+}{\partial\pos}  + \frac{\partial\pluckerTransform[k2]{k1}}{\partial\pos}\spatialVel[2]{k2}^+  &\text{(loop closure)}\\
&- \pluckerTransform[k2]{k1}\frac{\partial\spatialVel[2]{k2}^+}{\partial\pos},
\end{align}
where $\frac{\partial\!\spatialVel[1]{}^+}{\partial\pos}$, $\frac{\partial\!\spatialVel[2]{}^+}{\partial\pos}$ are the analytical derivatives of spatial velocities described in~\cite{carpentier2018analytical}, which are computed at the post-impact state $\state^+ = (\pos^+, \vel^+)$.
The additional term involving the derivative of the Plücker transform can be written as
\begin{align}\label{eq:pluckerVelDerivatives}
\frac{\partial\pluckerTransform[k2]{k1}}{\partial\pos}\spatialVel[2]{k2}^+ = -[\pluckerTransform[k2]{k1}\spatialVel[2]{k2}^+]_\times (\pluckerTransform[k2]{k1}\motionJac[2]{k2} - \motionJac[1]{k1}),
\end{align}
which follows from the identity $\frac{d\pluckerTransform[2]{1}}{dt} = [\pluckerTransform[2]{1}\spatialVel[2]{}-\spatialVel[1]{}]_{\times}\pluckerTransform[2]{1}$ (see~\cite{featherstone2014rigid}), where $[\cdot]_\times$ is the spatial cross-product operator (i.e., the \emph{small adjoint} $\mathrm{ad}\spatialVel[]{}$).
Similarly, the derivatives of the world-aligned bilateral contact are obtained as:
\begin{align}
\frac{\partial\residualVelContactLWA}{\partial\pos} = \lwaTransform[1]\frac{\partial\residualVelContact}{\partial\pos} + \frac{\partial\lwaTransform[1]}{\partial\pos}\pluckerTransform[1]{c}\,\frac{\partial\spatialVel[1]{}}{\partial\pos},
\end{align}
where the expression of the second term is derived in the same way as~\Cref{eq:pluckerVelDerivatives}.

\subsubsection{Contacts and closed-loop mechanisms within modes}
Differentiating the holonomic constraints a second time yields the acceleration-level conditions for bilateral contacts and closed-loop mechanisms, which are used in the forward dynamics phase of each mode. These conditions take the form:
\begin{align}\nonumber
\ddimplicitContact(\pos, \vel, \acc) &= \pluckerTransform[1]{c}\spatialAcc[1]{}, \quad &\text{(bilateral contact)} \\\nonumber
\ddimplicitLoop(\pos, \vel, \acc) &= \spatialAcc[1]{k1} - \pluckerTransform[k2]{k1} \spatialAcc[2]{k2} &\text{(loop closure)}\\\nonumber
&+ [\spatialVel[1]{k1}]_\times \pluckerTransform[k2]{k1} \spatialVel[2]{k2}, \quad \hspace{-0.3em}
\end{align}
with:\vspace{-1.8em}
\begin{align}\label{eq:acc_holonomic_constraints}
\spatialAcc[1]{k1} = \pluckerTransform[1]{k1}\spatialAcc[1]{}, \quad \spatialAcc[1]{k2} = \pluckerTransform[2]{k2}\spatialAcc[2]{},
\end{align}
where $\spatialAcc[1]{}$ and $\spatialAcc[2]{}$ are the spatial accelerations of the first and second bodies, expressed in their respective local coordinates, and the third term is derived similarly to~\Cref{eq:pluckerVelDerivatives}.
We define the total motion-constraint residual as:
\begin{align}
\ddimplicitConst{}(\pos, \vel, \acc) = \ddimplicitContact(\pos,\vel,\acc) - \biasAcc^*(\pos, \vel),
\end{align}
where the bias acceleration $\biasAcc^*(\pos,\vel)$ collects optional feedback correction in the constraint space.
Additionally, in the bilateral case to ease the enforcement of friction cones, we can project the constraint into a world-aligned frame, as described in~\Cref{eq:lwaTransform}, i.e.,
\begin{align}\label{eq:biasAccLWA}
\biasAccLWA(\pos, \vel) \;&=\; \lwaTransform[1](\pos)\,\biasAcc(\pos, \vel),
\end{align}
as its analytical derivatives include an additional rotation term, analogous to the velocity-level case, which we discussed in the previous section.

The configuration derivatives of the acceleration residuals are:
\begin{align}\label{eq:accDerivatives}\nonumber
\frac{\partial\residualAccContact}{\partial\pos} \;&=\; \pluckerTransform[1]{c}\,\frac{\partial\spatialAcc[1]{}}{\partial\pos} - \frac{\partial\biasAcc[{b_c}]^*}{\partial\pos}, \quad\hspace{4.7em} \text{(bilateral contact)} \\\nonumber
\frac{\partial\residualAccLoop}{\partial\pos} \;&=\; 
\frac{\partial\spatialAcc[1]{k1}}{\partial\pos}
- \frac{\partial\biasAcc[{b_k}]^*}{\partial\pos}
- \pluckerTransform[k2]{k1}\,\frac{\partial\spatialAcc[2]{k2}}{\partial\pos}
\quad\hspace{0.7em} \text{(loop closure)} \\\nonumber  
&- \frac{\partial\pluckerTransform[k2]{k1}}{\partial\pos} \spatialAcc[2]{k2} + \left[\frac{\partial\spatialVel[1]{k1}}{\partial\pos}\right]_\times \pluckerTransform[k2]{k1} \spatialVel[2]{k2}
\\  
&+ [\spatialVel[1]{k1}]_\times \left( \pluckerTransform[k2]{k1} \frac{\partial\spatialVel[2]{k2}}{\partial\pos} + \frac{\partial\pluckerTransform[k2]{k1}}{\partial\pos} \spatialVel[2]{k2} \right),
\end{align}
where $\frac{\partial\!\spatialAcc[1]{}}{\partial\pos}$ and $\frac{\partial\!\spatialAcc[2]{}}{\partial\pos}$ are the analytical derivatives of spatial accelerations~\cite{carpentier2018analytical} expressed in their respective frames, $\frac{\partial\biasAcc[{b_c}]^*}{\partial\pos}$ and $\frac{\partial\biasAcc[{b_k}]^*}{\partial\pos}$ are the partial derivatives of the corrective terms, and the extra terms arise from the cross-product interaction of spatial velocities or transformation of these partial derivatives.
The derivative of the Plücker transform follows the same identity as in the velocity case.

Similarly, the velocity derivatives of the acceleration residual are:
\begin{align}\label{eq:accDerivativesVel}\nonumber
\frac{\partial\residualAccContact}{\partial\vel} \;&=\; \pluckerTransform[1]{c}\,\frac{\partial\spatialAcc[1]{}}{\partial\vel} - \frac{\partial\biasAcc[{b_c}]^*}{\partial\vel}, \quad\hspace{4.7em} \text{(bilateral contact)} \\\nonumber
\frac{\partial\residualAccLoop}{\partial\vel} \;&=\;
\frac{\partial\!\spatialAcc[1]{k1}}{\partial\vel}
- \frac{\partial\biasAcc[{b_k}]^*}{\partial\vel}
- \pluckerTransform[k2]{k1} \frac{\partial\spatialAcc[2]{k2}}{\partial\vel} \quad\hspace{0.7em} \text{(loop closure)}
\\ +& \left[\frac{\partial\spatialVel[1]{k1}}{\partial\vel}\right]_\times \pluckerTransform[k2]{k1} \spatialVel[2]{k2} + [\spatialVel[1]{k1}]_\times \pluckerTransform[k2]{k1} \frac{\partial\spatialVel[2]{k2}}{\partial\vel}.
\end{align}

In addition to the motion-constraint derivatives, we also need to differentiate the generalized forces induced by the constraints.
For each constraint $i$ 
the corresponding generalized force is
$
  \eff[i](\pos,\constraintForce_i)
  \;=\;
  \motionJac[i]{}(\pos)^\transpose \constraintForce_i
$,
and the total constraint contribution is
$\eff = \sum_i \eff[i]$. Differentiating with respect to the configuration
yields
\begin{equation}
  \frac{\partial\eff}{\partial \pos}
  \;=\;
  \sum_i \left(
    \frac{\partial \motionJac[i]{}^\transpose}{\partial \pos}\,\constraintForce_i
    \;+\;
    \motionJac[i]{}^\transpose \frac{\partial \constraintForce_i}{\partial \pos}
  \right).
\end{equation}
When each wrench $\constraintForce_i$ is expressed in the same joint-local
frame as $\motionJac[i]{}(\pos)$ (e.g., \emph{local} contacts), the term
$\partial \motionJac[i]{}^\transpose / \partial \pos$ is exactly the one already
handled by the standard \gls{rnea} derivatives.  In that case,
no additional contribution to $\partial\eff/\partial\pos$ is required.

Bilateral contacts in the world frame $\worldFrame$ and closed-loop mechanisms introduce an extra geometric-stiffness term because their wrenches are parameterized in frames that move with the configuration (see~\Cref{eq:lwaJac,eq:dholonomic_constraints}).  Let $\constraintForce_i$ be expressed in a frame $\spatialFrame[c]$ related to a joint frame $\spatialFrame[j]$ by a Plücker transform
$\pluckerTransform[c]{j}(\pos)$.  The generalized force at joint
$\spatialFrame[j]$ reads
\begin{equation}
  \eff[c](\pos,\constraintForce_c)
  \;=\;
  \motionJac[i]{}(\pos)^\transpose
  \pluckerTransform[c]{j}(\pos)^\transpose \constraintForce_i,
\end{equation}
so that differentiating the transform yields
\begin{equation}
  \frac{\partial}{\partial\pos}
  \Big( \pluckerTransform[c]{j}(\pos)^\transpose \constraintForce_i \Big)
  \;=\;
  - \big[ \pluckerTransform[c]{j}(\pos)^\transpose \constraintForce_i \big]_\times^\ast
    \,\motionJac[c]{}(\pos),
\end{equation}
This induces an additional contribution of the form
\begin{equation}
  \frac{\partial \eff[i]}{\partial\pos}
  \;\supset\;
  -\,\motionJac[c]{}(\pos)^\transpose
  \big[ \pluckerTransform[c]{j}(\pos)^\transpose \constraintForce_i \big]_\times^\ast
  \,\motionJac[c]{}(\pos),
\end{equation}
which needs to be accounted for the bilateral contacts expressed in the world frame $\worldFrame$ and the force acting on body $2$ in loop closure constraints.

When the contact is expressed in world-aligned coordinates the accelaration bias is mapped as in~\Cref{eq:biasAccLWA},
and the derivatives follow the same rotation-skew correction used at velocity level:
\begin{align}
\frac{\partial \biasAccLWA^{*}}{\partial \pos}
&= \lwaTransform[1]\frac{\partial \biasAcc^{*}}{\partial \pos}-
\begin{bmatrix}
\spatialAcc[1]{}^{\omega}\times\rotMatrix[1]{\worldFrame}\contactJac^{\omega}\\[2pt]
\spatialAcc[1]{}^{v}\times\rotMatrix[1]{\worldFrame}\contactJac^{\omega}
\end{bmatrix},
\\\nonumber
\frac{\partial \biasAccLWA^{*}}{\partial \vel}
&= \lwaTransform[1]\,\frac{\partial \biasAcc^{*}}{\partial \vel},
\end{align}
with $\spatialAcc[1]{}^{v}$ and $\spatialAcc[1]{}^{\omega}$ the contact linear/angular accelerations in world coordinates.

%% file: chapters/disturbed_dynamics.tex
\section{Hybrid Dynamics under External Disturbances}\label{sec:disturbance_under_implicit_const}

To avoid relying on known state trajectories in~\gls{sysid}, we must consider the state evolution as a \emph{stochastic} process with additive Gaussian noise~\cite{welch1995introduction}:
\begin{align}\label{eq:int}
\state'=\Integrator(\state,\acc) \oplus\ucertain, \quad \ucertain \sim \mathcal{N}(\zeroVec,\ucertainCov),
\end{align}
where $\state=(\pos,\vel)\in\stateManif$ denotes the current state, $\Integrator:\stateManif\times\stateTManif\to\stateManif$ is the numerical integrator that maps the current state and accelerations to the next state $\state'$, and $\ucertainCov$ is the covariance of the process noise $\ucertain\in\R^\nq$.
To ensure consistency, the process noise must be defined in a way that respects implicit constraints, as detailed in the next section.

\subsection{Disturbance with Implicit Motion Constraints}
When implicit motion constraints are present, admissible accelerations belong to the tangent space of the implicit constraint manifold $\implConstTManif = \{\pos\in\confManif, \vel,\acc\in\confTManif \mid \contactJac(\pos)\,\acc=\biasAcc(\pos,\vel)\}$.
To preserve consistency with the system's connectivity graph, process noise must be injected only along feasible directions.
We achieve this by projecting any process vector in $\R^\nq$ onto the tangent space of the constraints $\implConstTManif$.
This ensures that the resulting uncertainty remains consistent with respect the system's kinematic structure.

The projection is defined by a matrix $\mathbf{N}(\pos)\in\R^{\nq\times\nq}$ that spans the kernel of the constraint Jacobian $\motionJac{}(\pos)$ and captures the feasible motion directions.
From~\Cref{eq:kkt_acc}, the constrained-acceleration update is given by
\begin{align}\label{eq:proj_acc}
&\acc = \acc[\text{free}] + \massMatrix^{-1} \motionJac{}^\transpose(\motionJac{} \massMatrix^{-1} \motionJac{}^\transpose)^{-1}(\biasAcc - \motionJac{} \acc[\text{free}]),
\end{align}
from which we identify the \emph{dynamically consistent projector} onto the constraint nullspace as
\begin{align}
\label{eq:dyn_proj}
\mathbf{N}(\pos) \coloneqq\ \mathbf{I} - \massMatrix^{-1}\motionJac{}^\transpose(\motionJac{} \massMatrix^{-1}\motionJac{}^\transpose)^{-1}\motionJac{}.
\end{align}
This projector enforces energy-momentum consistency, which explains the appearance of the mass matrix as a weighting term.
For comparison, a purely kinematic projector onto the constraint nullspace can be written as: $\mathbf{N}(\pos) = \mathbf{I} - \motionJac{}^\transpose\,(\motionJac{} \motionJac{}^\transpose)^{-1}\motionJac{}$,
which is computationally cheaper and independent of inertial parameters.
However, the dynamically consistent projector guarantees physically meaningful accelerations and forces, making it essential for \gls{sysid} and Bayesian estimation.
Concretely, we project the process noise as:
\begin{align}
\label{eq:noise_proj}
\ucertainProj \;=\; 
\underbrace{\begin{bmatrix}
\mathbf{N}(\pos) & \\ 
 & \mathbf{N}(\pos)
\end{bmatrix}}_{\mathbf{P}(\pos)}
\begin{bmatrix}
\ucertain[\pos] \\ \ucertain[\vel]
\end{bmatrix},
\quad
\ucertainProjCov \;=\; \mathbf{P}(\pos)\,\ucertainCov\,\mathbf{P}(\pos)^\transpose,
\end{align}
where $\ucertain[\pos]$ and $\ucertain[\vel]$ are the position and velocity components of $\ucertain$, respectively; $\ucertainProj$ and $\ucertainProjCov$ denote the projected noise and covariance, respectively, consistent with the implicit motion constraints.

\subsection{Analytical Derivatives of the Dynamically Consistent Projector}
The projector introduced in~\Cref{eq:dyn_proj} can be seen as the result of applying Gauss's principle of least constraint to the process noise, i.e.,
\begin{equation}\label{eq:gauss_noise}
\begin{aligned}
\min_{\ucertainProj[\arbitraryVar]} \quad
    & \tfrac{1}{2}
      \bigl\|
        \ucertainProj
        - \ucertain
      \bigr\|^2_{\massMatrix(\pos)} \\
\text{subject to} \quad
    & \motionJac[]{}(\pos)\,\ucertainProj
      = \zeroVec,
\end{aligned}
\end{equation}
where the $\arbitraryVar$ subscript refers to either $\pos$ or $\vel$.
The corresponding~\gls{kkt} conditions of~\Cref{eq:gauss_noise} are:
\begin{align}
\label{eq:kkt_noise}
\begin{bmatrix}
\ucertainProj  \\ -\boldsymbol{\zeta}
\end{bmatrix}
=
\begin{bmatrix}
\massMatrix(\pos) & \motionJac[]{}(\pos)^\transpose \\
\motionJac[]{}(\pos) &
\end{bmatrix}^{-1}
\begin{bmatrix}
\massMatrix(\pos)\ucertain \\ \zeroVec
\end{bmatrix},
\end{align}
whose solution yields~\Cref{eq:dyn_proj}, where $\boldsymbol{\zeta}$ denotes the Lagrange multiplier associated with the motion constraint. 
Similarly to the previous section, we define the corresponding residuals as follows:
\begin{align}
\mathbf{r} =
\begin{bmatrix}
\mathbf{r}_{w} \\[2pt] \mathbf{r}_{\zeta}
\end{bmatrix}
=
\begin{bmatrix}
\massMatrix(\pos)(\ucertainProj - \ucertain) - \motionJac{}(\pos)^\transpose\boldsymbol{\zeta} \\[2pt]
\motionJac{}(\pos)\,\ucertainProj
\end{bmatrix}.
\end{align}

Analogously to \Cref{eq:derivative_kkt_vel}, the forward-mode derivatives of the Gauss projection~\gls{kkt} system are
\begin{align}
\label{eq:derivative_kkt_gauss}
\begin{bmatrix}
\frac{\partial \ucertainProj[\arbitraryVar]}{\partial \arbitraryVar} \\[2pt]
-\frac{\partial \boldsymbol{\zeta}}{\partial \arbitraryVar}
\end{bmatrix}
=
-
\begin{bmatrix}
\massMatrix(\pos) & \motionJac{}(\pos)^\transpose \\
\motionJac{}(\pos) &
\end{bmatrix}^{-1}
\begin{bmatrix}
\frac{\partial \mathbf{r}_w(\pos)}{\partial \arbitraryVar} \\
\frac{\partial \mathbf{r}_{\zeta}(\pos)}{\partial \arbitraryVar}
\end{bmatrix},
\end{align}
where the residual derivatives are similarly computed as in \Cref{eq:derivative_impulse} with $\postVel$, $\preVel$ replaced by $\ucertainProj$, $\ucertain$, respectively.
Additionally, we reuse the factorization of~\Cref{eq:kkt_acc} for enhancing computational efficiency.
Next, we describe the dynamic parameters typically considered in \gls{sysid}.

%% file: chapters/inertial_actuation.tex
\section{Inertial and Actuation Parameters}\label{sec:dyn_inertia}
The dynamics in~\Cref{eq:dyn,eq:kkt_acc,eq:kkt_vel_reset} can be parameterized by the inertial parameters $\inertialParams\in\R^{10\nb}$ and actuation parameters $\frictionParams\in\R^{\nactparams}$.
For efficiency, both the dynamics and their analytical derivatives (see~\Cref{sec:background,sec:derivatives_of_hybrid_dynamics_with_closed_loops}) are computed using the~\gls{imm}~\cite{featherstone2014rigid}.
Consistency with the~\gls{imm} is essential, as its Cholesky-based factorization of the inertia matrix implicitly constrains the set of physically admissible inertial parameters.

\subsection{Inertial Parameters and Physical Consistency}

\begin{figure}[t]\centering
\href{\video&t=64}{\includegraphics[width=1.0\linewidth]{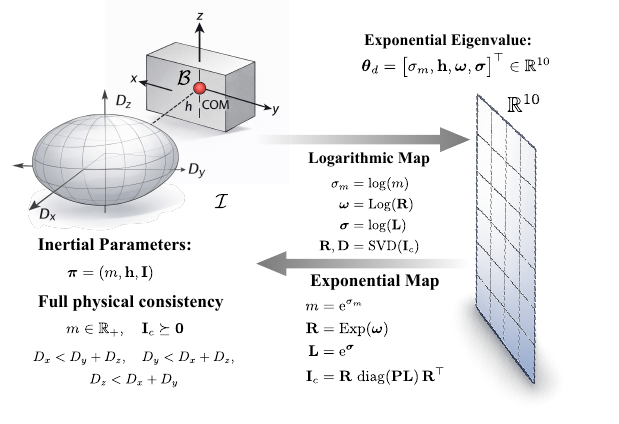}}
    \caption{ Exponential--eigenvalue (EE) parametrization of a rigid body’s inertial parameters. The physical quantities $\boldsymbol{\pi} = (m, \mathbf{h}, \mathrm{vech}(\mathbf{I}))$ (mass, first moment, and inertia) are illustrated by the body frame $\mathcal{B}$, CoM, and inertia ellipsoid with principal moments $(D_x,D_y,D_z)$.
    The unconstrained coordinates $\dynparams\in\mathbb{R}^{10}$ (shown as a mesh) are mapped to a physically consistent inertia via the exponential map, guaranteeing $m \ge 0$, $\rotInertiaBarycenter \succeq \zeroVec$, and triangle inequalities on the principal moments; the inverse (logarithmic) map recovers $\dynparams$ from $\boldsymbol{\pi}$.}
    \label{fig:EE_inertial}
\end{figure}
The inertial parameters of a rigid body $i$ are represented by the vector $\inertialParams[i]\in \R^{10}$:
\begin{align}
\label{eq:inertia_param}
\inertialParams[i]\;=\;
\begin{bmatrix}
\mass_i & \lever_i^\transpose & \mathrm{vech}(\rotInertia_i)^\transpose
\end{bmatrix}^\transpose \in \mathbb{R}^{10},
\end{align}
where $\mass_i\in\R_+$ is the body mass, $\lever_i=\mass_i\,\barycenter_i$ is the first mass moment with~\gls{com} $\barycenter_i\in\mathbb{R}^3$, and rotational inertia $\rotInertia_i$ expressed in the body frame.
The rotational inertia at the~\gls{com} is obtained as
$\rotInertiaBarycenter = \rotInertia - \tfrac{1}{\mass}[\lever]_{\times} [\lever]_{\times}^\transpose$, with $[\cdot]_{\times}$ denoting the skew-symmetric operator.

Stacking all bodies’ parameters yields $\inertialParams=(\inertialParams[i])_{i=0}^{n_b}\in\mathbb{R}^{10\nb}$.
Following~\cite{atkeson1986estimation}, the generalized torques can be then expressed as an affine function of $\inertialParams$, i.e.,
\begin{gather}\label{eq:NE_param} 
\eff(\ctrlMeas) = \regInertialMatrix(\pos,\vel,\acc)\inertialParams,
\end{gather}
where $\regInertialMatrix:\confManif\times\confTManif\times\confTManif\to\R^{\nq\times 10\nb}$ denotes the joint-torque regressor matrix.

The inertial parameters $\inertialParams[i]$ are \textit{fully physically consistent} $\inertialParams[i]\in\inertialManif$, if they satisfy:
\begin{gather}\label{eq:physical_consistent_cond}
\begin{aligned}
\mass\in\R_+, \quad &\rotInertiaBarycenter \succeq \zeroVec, \\
D_x < D_y + D_z, \quad D_y < D_x +& D_z, \quad D_z < D_x + D_y.
\end{aligned}
\end{gather}
where $\eigInertia=\mathrm{diag}(D_x, D_y, D_z)\in\R^{3\times 3}$ are the principal moments of inertia, obtained from $\rotInertiaBarycenter=\rotMatrix[]{}\eigInertia\rotMatrix[]{}^\transpose$ with $\rotMatrix[]{}\in\SO[3]$.
The condition $\rotInertiaBarycenter \succeq \zeroVec$ ensures that the principal components of inertia are nonnegative, while the triangle inequalities guarantee the \textit{physical realization} of the second moments of mass, i.e., consistency with a positive mass distribution~\cite{traversaroidentification}.
These inequalities can be compactly written as $\secMomentMass\in\R^3_+$, where $\secMomentMass=(L_x, L_y, L_z)$ denotes the second moment of inertia, related to the principal moments by
\begin{gather}
\eigInertia = \mathrm{diag}\left(\mathbf{P}\secMomentMass\right), \,\,\text{with}\,\,\mathbf{P}=\begin{bmatrix}0 & 1 & 1\\ 1 & 0 & 1\\1 & 1 & 0\end{bmatrix}.
\end{gather}

To enforce these conditions by construction, we adopt the \emph{Exponential--Eigenvalue} (EE) parametrization~\cite{martinez2025multi}:
\begin{align}\label{eq:exp_eig}
\setlength\arraycolsep{2pt}
\mass &= \exp(\sigma_\mass), \quad
\rotMatrix{} = \mathrm{Exp}(\rotMatrixParam),\\\nonumber
\secMomentMass &=
\begin{bmatrix}
\exp(\sigma_x) & \exp(\sigma_y) & \exp(\sigma_z)
\end{bmatrix}^\transpose,
\end{align}
where $\mathrm{Exp}$ is the exponential map of the $\SO[3]$.
This parametrization enforces positivity by construction, yields smooth derivatives, and admits a Lie algebra representation suitable for interpretable priors (see~\Cref{fig:EE_inertial}).
The inertial parameters are finally reparametrized as:
\begin{gather}\label{eq:eeval_param}
\setlength\arraycolsep{2pt}
\dynparams=
\begin{bmatrix}
\sigma_m& h_x& h_y& h_z& \omega_x& \omega_y& \omega_z& \sigma_x& \sigma_y& \sigma_z
\end{bmatrix}^\transpose \in \mathbb{R}^{10},
\end{gather}
where $(\sigma_m,\sigma_x,\sigma_y,\sigma_z)$ ensures positivity through exponentiation, $\rotMatrixParam=(\omega_x,\omega_y,\omega_z)$ are the rotation parameters in $\mathfrak{so}(3)$, and $(h_x,h_y,h_z)$ define the first mass moment.

\subsubsection{Derivatives of the inertial parametrization}\label{sec:inertial_param_derivatives}
The derivatives $\partial\inertialParams / \partial\dynparams$ are obtained in closed form by grouping contributions per parameter block.
The mass and first mass moment give
\begin{align}
  \frac{\partial \mass}{\partial \sigma_m} &= \mass, &
  \frac{\partial \lever}{\partial (h_x,h_y,h_z)} &= \eyeMatrix[3].
\end{align}
For the rotational inertia, we have
\begin{align}
  \frac{\partial \rotInertia}{\partial \sigma_m}
  &= \frac{\partial \rotInertia}{\partial \mass}\frac{\partial \mass}{\partial \sigma_m}
   = -\frac{1}{\mass}[\lever]_{\times} [\lever]_{\times}^\transpose,\\\nonumber
  \frac{\partial \rotInertia}{\partial h_k}
  &= \frac{1}{\mass}\frac{\partial ([\lever]_{\times} [\lever]_{\times}^\transpose)}{\partial h_k}, \\\nonumber
  &= \frac{1}{\mass}\Bigl(2 h_k \eyeMatrix[3] - \mathbf{e}_k \lever^\transpose - \lever \mathbf{e}_k^\transpose\Bigr),
   \qquad k\in\{x,y,z\},
\end{align}
where $\mathbf{e}_x,\mathbf{e}_y,\mathbf{e}_z\in\R^3$ denote the canonical basis vectors.
For the rotation parameters, if we denote $\mathbf{J}_\ell(\rotMatrixParam)$ as the left Jacobian of $\SO[3]$ and $\mathbf{u}_i = \mathbf{J}_\ell(\rotMatrixParam)\,\mathbf{e}_i$, we have
\begin{align}
  \frac{\partial \rotMatrix{}}{\partial \omega_i}
  &= \rotMatrix{}[\mathbf{u}_i]_\times,
  \qquad i\in\{x,y,z\}.
\end{align}
Since $\rotInertiaBarycenter = \rotMatrix{}\eigInertia\rotMatrix{}^\transpose$, it follows that
\begin{align}
  \frac{\partial \rotInertia}{\partial \omega_i}
  &= \frac{\partial \rotInertiaBarycenter}{\partial \omega_i}
   = \rotMatrix{}[\mathbf{u}_i]_\times \eigInertia \rotMatrix{}^\transpose
    + \rotMatrix{}\eigInertia [\mathbf{u}_i]_\times^\transpose \rotMatrix{}^\transpose,
    \qquad i\in\{x,y,z\}.
\end{align}

For the exponential eigenvalues, from~\Cref{eq:exp_eig} we have
$\secMomentMass = (L_x,L_y,L_z)$ with $L_\alpha = \exp(\sigma_\alpha)$ and
$\eigInertia = \mathrm{diag}(\mathbf{P}\secMomentMass)$.
Hence
\begin{align}
  \frac{\partial \secMomentMass}{\partial \sigma_x}
  &= [L_x,0,0]^\transpose, &
  \frac{\partial \secMomentMass}{\partial \sigma_y}
  &= [0,L_y,0]^\transpose, &
  \frac{\partial \secMomentMass}{\partial \sigma_z}
  &= [0,0,L_z]^\transpose,
\end{align}
which yields
\begin{align}
  \frac{\partial \eigInertia}{\partial \sigma_x}
  &= \mathrm{diag}(0,L_x,L_x), \quad
  \frac{\partial \eigInertia}{\partial \sigma_y}
  = \mathrm{diag}(L_y,0,L_y), \\\nonumber
  \frac{\partial \eigInertia}{\partial \sigma_z}
  &= \mathrm{diag}(L_z,L_z,0),
\end{align}
and therefore
\begin{align}
  \frac{\partial \rotInertia}{\partial \sigma_\alpha}
  &= \rotMatrix{}\,\frac{\partial \eigInertia}{\partial \sigma_\alpha}\,\rotMatrix{}^\transpose,
  \qquad \alpha\in\{x,y,z\}.
\end{align}

The Jacobian $\partial\inertialParams / \partial\dynparams \in \R^{10\times 10}$ is finally obtained by stacking,
for each parameter in $\dynparams$, the derivative of $\mass$, $\lever$, and each of the six independent entries of rotational inertia.

\subsection{Actuation and Joint-Friction Parameters}
The mapping from commanded inputs to generalized torques is defined as
\begin{align}
\label{eq:general_actuation}
\eff(\pos,\vel;\frictionParams | \ctrlMeas)
\;=\;
\mathbf{S}(\frictionParams)\,\Big(\ctrlMeas - \friction(\vel;\frictionParams)\Big),
\end{align}
where $\mathbf{S}(\frictionParams)$ is the parameterized selection matrix mapping actuated joints to generalized coordinates, $\ctrlMeas$ denotes the commanded effort (e.g., motor torque), and $\friction(\vel;\frictionParams)$ represents the joint friction coefficients.
In adition to identifying the friction model, parametrizing the elements of $\mathbf{S}$ enables identification of actuation coefficients such as gear ratios or momentum-thrust coupling in systems like quadrotors or underwater vehicles.

\begin{figure}[b]\centering
    \href{\video&t=149}{\includegraphics[width=\linewidth]{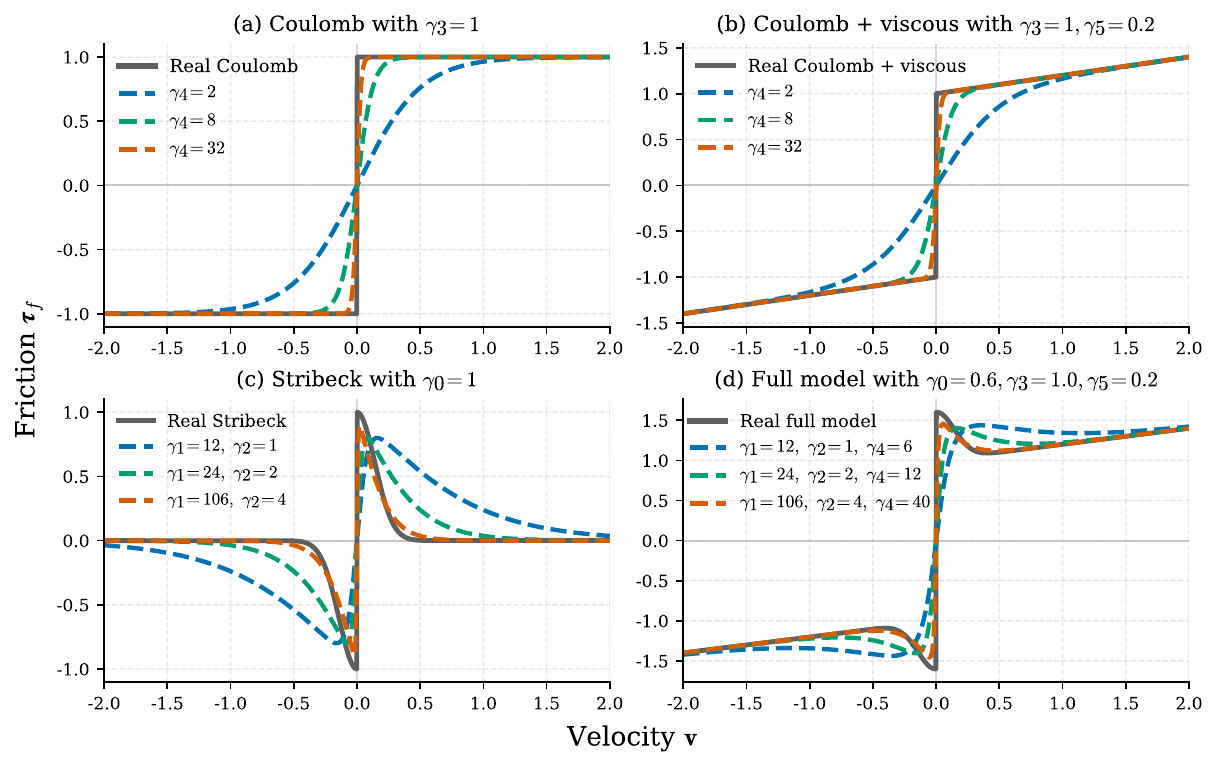}}
    \caption{
        Visualization of friction models evaluated at different smoothing coefficients. 
        The friction model remains smooth and differentiable while capturing the Coulomb, viscous, and Stribeck effects.
    }
    \label{fig:friction_models}
\end{figure}

Unlike classical approaches, we explicitly model the full nonlinear joint friction using the smooth and differentiable formulation proposed in~\cite{makkar2005new}:
\begin{align}
\label{eq:friction}
\tau_f(v,\frictionParams)
\;=\; &- \frictionParams[0]\!\left(\tanh(\frictionParams[1]v) - \tanh(\frictionParams[2]v)\right) \\\nonumber
& + \frictionParams[3]\tanh(\frictionParams[4]v) + \frictionParams[5]v,
\end{align}
where $\tau_f$, $v$ denote the joint friction and velocity, respectively.
This compact model captures stiction, Coulomb, viscous, and Stribeck friction within a single smooth formulation.
Its analytical derivatives make it particularly suitable for gradient-based system identification.
\Cref{fig:friction_models} illustrates the contributions of each friction components.

Despite its expressiveness, the model in~\Cref{eq:friction} does not explicitly enforce physical consistency, and may therefore allow parameter values under which friction appears to generate, rather than dissipate energy.
To ensure \textit{physical plausibility}, we impose constraints that guarantee friction always resists motion:
\begin{align}\label{eq:friction_const}
&\frictionParams[0], \frictionParams[1], \frictionParams[3], \frictionParams[4], \frictionParams[5] \ge 0,
\qquad
\frictionParams[1] \ge \frictionParams[2],
\end{align}
where positivity is enforced via exponential maps:
\begin{align}
&\frictionParams[0] = \mathrm{exp}(\mu_0),\quad \frictionParams[1] = \mathrm{exp}(\mu_1),\quad \frictionParams[3] = \mathrm{exp}(\mu_3),\\\nonumber
&\frictionParams[4] = \mathrm{exp}(\mu_4),\quad \frictionParams[5] = \mathrm{exp}(\mu_5),
\end{align}
with $\actparams =\begin{bmatrix}\mu_0& \mu_1& \mu_2& \mu_3& \mu_4& \mu_5\end{bmatrix}^\transpose \in \mathbb{R}^{6}$ are joint-friction parameters, and the inequality $\frictionParams[1]\!\ge\!\frictionParams[2]$ is handled using the stagewise optimizer described in~\Cref{sec:parametrized_riccati}.

\subsubsection{Derivatives of the joint friction}
We require the derivatives of the friction torque
$\tau_f(v,\frictionParams)$ with respect to the joint velocity $v$ and the
actuation parameters $\actparams$.
The velocity derivative is
\begin{align}\nonumber
  \frac{\partial \tau_f}{\partial v}
  = -\gamma_0 \gamma_1 c_1(v)
    + \gamma_0 \gamma_2 c_2(v)
    + \gamma_3 \gamma_4 c_4(v)
    + \gamma_5.
\end{align}
with $c_i(v) \coloneqq 1 - \tanh^2(\gamma_i v)$ for $i\in\{1,2,4\}$.
The derivatives with respect to the joint-friction parameters are
\begin{align}\nonumber
  \frac{\partial \tau_f}{\partial \gamma_0}
  &= -\Big(\tanh(\gamma_1 v) - \tanh(\gamma_2 v)\Big), &
  \frac{\partial \tau_f}{\partial \gamma_1}
  &= -\gamma_0\,v\,c_1(v),\\\nonumber
  \frac{\partial \tau_f}{\partial \gamma_2}
  &= \gamma_0\,v\,c_2(v),  &
  \frac{\partial \tau_f}{\partial \gamma_3}
  &= \tanh(\gamma_4 v),\\
  \frac{\partial \tau_f}{\partial \gamma_4}
  &= \gamma_3\,v\,c_4(v), &
  \frac{\partial \tau_f}{\partial \gamma_5}
  &= v.
\end{align}
To account for the use of exponential mappings to enforce the constraints in~\Cref{eq:friction_const}, the Jacobian with respect to $\actparams$ is obtained via the chain rule:
\begin{align}
  \frac{\partial \tau_f}{\partial \actparams}
  = \frac{\partial \tau_f}{\partial \frictionParams}
    \,\frac{\partial \frictionParams}{\partial \actparams}.
\end{align}
Finally, the derivatives of the generalized \emph{joint-friction effort} are
\begin{align}
  \frac{\partial \eff}{\partial \vel}
  &= -\,\mathbf{S}(\actparams)\,\frac{\partial \friction(\vel;\actparams)}{\partial \vel},\\\nonumber
  \frac{\partial \eff}{\partial \actparams}
  &= \frac{\partial \mathbf{S}}{\partial \actparams}
     \bigl(\ctrlMeas - \friction(\vel;\actparams)\bigr)
     - \mathbf{S}(\actparams)\,
       \frac{\partial \friction(\vel;\actparams)}{\partial \actparams},
\end{align}
where $\partial\friction/\partial\vel$ and
$\partial\friction/\partial\actparams$ are assembled from the scalar expressions above for each actuated joint.

Estimating joint friction jointly with inertia parameters is challenging; we therefore adopt an energy-based approach, described in the next section.

%% file: chapters/energy.tex
\section{Energy-based observations}\label{sec:energy}
The principle of energy conservation can be leveraged by introducing energy-based observations in \gls{sysid}.
This principle states that the total mechanical energy of a system remains constant in the absence of external work.
However, since the total energy cannot be directly measured, nor accurately computed without known physical parameters, it is more practical to express this relationship in terms of actuation power~\cite{gautier1997dynamic}:
\begin{align}\label{eq:input_energy}
\underbrace{\int_{a}^{b} \vel^\transpose\eff(\pos,\vel;\actparams | \ctrlMeas) \mathrm{d}t}_{\text{input energy}} = 
\underbrace{\MechanicalEnergy_{b}(\state;\dynparams) - \MechanicalEnergy_{a}(\state;\dynparams | \ctrlMeas)}_{\text{change in mechanical energy}},
\end{align}
where $\MechanicalEnergy = \KineticEnergy + \PotentialEnergy \in \R_+$ denotes the total mechanical energy of the system, composed of kinetic $\KineticEnergy$ and potential $\PotentialEnergy$ terms, and the subscripts $a$ and $b$ correspond to the integration limits or discrete timesteps.
Therefore, we define an energy observation model as:
\begin{align}
\label{eq:input_energy_res}
\obsFunc[E](\state;\params\vert\ctrlMeas) &= \MechanicalEnergy'(\state';\dynparams)
- \MechanicalEnergy(\state;\dynparams)
+ \MechanicalEnergy_f(\vel;\actparams) \\\nonumber
\end{align}
with
\begin{subequations}
\begin{align}\nonumber
&\state' = \Integrator(\state,\acc;\params \mid \ctrlMeas), \quad
\obsMeas[E] = \int_{t_k}^{t_{k+1}} \hat{\vel}^\transpose \ctrlMeas \, dt, \\\nonumber
&\MechanicalEnergy_f(\vel;\actparams) = \int_{t_k}^{t_{k+1}} \vel^\transpose \eff[f](\vel;\actparams) \, dt,
\end{align}
\end{subequations}
where $\obsMeas[E]\in\R_+$ is the observed input power, obtained from the joint observations, and $\MechanicalEnergy_f$ is the dissipative energy due to joint-friction effects computed from the actuation model. 
This observation enables the identification of trajectories whose rate of change of mechanical energy is consistent not only with the supplied power, but also with a joint-friction model that accurately captures dissipative effects.

We apply the chain rule to derive the analytical derivatives of~\Cref{eq:input_energy_res}, i.e.,
\begin{align}\label{eq:dEnergy_dp}
\frac{\partial \obsFunc[E]}{\partial\state} =&
\frac{\partial\MechanicalEnergy'}{\partial\state} \dynFunc[\state] -
\frac{\partial\MechanicalEnergy}{\partial\state} +
\begin{bmatrix}
\zeroVec^\transpose\\
\frac{\partial \MechanicalEnergy_f}{\partial\vel}
\end{bmatrix}, \qquad
\frac{\partial \obsFunc[E]}{\partial\generalCtrl} =
\frac{\partial\MechanicalEnergy'}{\partial\state} \dynFunc[\generalCtrl],\\
\frac{\partial \obsFunc[E]}{\partial\params} =&
\frac{\partial\MechanicalEnergy'}{\partial\state} \dynFunc[\params] +
(\frac{\partial\MechanicalEnergy'}{\partial\inertialParams} -\frac{\partial\MechanicalEnergy}{\partial\inertialParams}) \frac{\partial\inertialParams}{\partial\params} + \frac{\partial \MechanicalEnergy_f}{\partial\params},
\end{align}
with:\vspace{-0.8em}
\begin{align}\nonumber
\frac{\partial\MechanicalEnergy}{\partial\state} = &\begin{bmatrix}
\frac{\partial \KineticEnergy}{\partial\pos} + \frac{\partial \PotentialEnergy}{\partial\pos}\\
\frac{\partial \KineticEnergy}{\partial\vel}
\end{bmatrix}, \quad
\frac{\partial\MechanicalEnergy}{\partial\inertialParams} =  \begin{bmatrix}
\frac{\partial \KineticEnergy}{\partial\inertialParams} + \frac{\partial \PotentialEnergy}{\partial\inertialParams}
\end{bmatrix},\nonumber
\end{align}
where $\dynFunc[\state], \dynFunc[\generalCtrl], \dynFunc[\params]$ are the Jacobians of the hybrid dynamics $\left(\state' = \Integrator(\state,\acc;\params \mid \ctrlMeas)\right)$ and $\frac{\partial\inertialParams}{\partial\params}$ is the derivative of the inertial parametrization, which are described in~\Cref{sec:derivatives_of_hybrid_dynamics_with_closed_loops} and~\Cref{sec:inertial_param_derivatives}, respectively.
Instead, to compute the derivatives of the kinetic and potential energies, we first introduce the corresponding energy regressors~\cite{gautier1997dynamic}.

\begin{figure}[t]\centering
    \href{\video&t=88}{\includegraphics[width=1.0\linewidth]{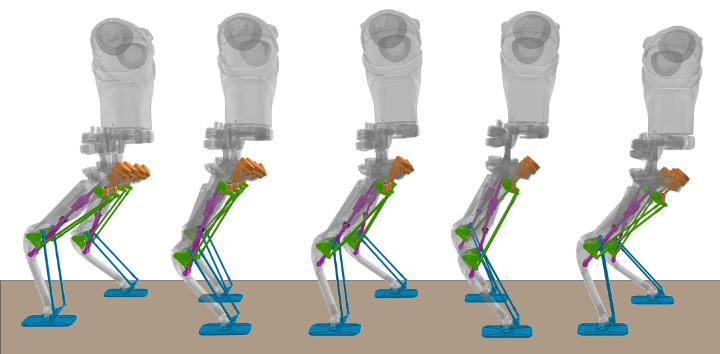}}
    \caption{
        Sequential snapshots of the Kangaroo robot executing multiple walking gaits. 
        Key leg components are highlighted: ankle differential drives (blue) transmitting through a second pair of differentials to the knee (green), then to hip-mounted actuators (orange), with a prismatic knee linkage forming an additional closure (magenta). 
        The presence of \emph{three loop closures in series} makes the mechanism highly constrained and the estimation/identification problem particularly challenging.
    }
    \label{fig:kangaroo_seq}
\end{figure}

\subsection{Energy Regressors}
The mechanical energy of a rigid-body system is defined as sum of its kinetic and potential energy, where
\begin{align}\label{eq:kinE}
\KineticEnergy(\pos, \vel; \inertialParams) =& \frac{1}{2} \vel^\transpose \massMatrix(\pos;\inertialParams) \vel \\
=& \frac{1}{2} \sum_{k=0}^{\nb} \boldsymbol{v}_k(\pos, \vel)^\transpose \boldsymbol{I}_k(\inertialParams) \boldsymbol{v}_k(\pos, \vel)\nonumber
\end{align}
is the kinetic energy, $\boldsymbol{v}_k$ is the spatial velocity, and $\boldsymbol{I}_k$ the spatial inertia of the $k$-th body~\cite{featherstone2014rigid}. 
The potential energy is defined as:
\begin{align}\label{eq:potE}
\PotentialEnergy(\pos;\inertialParams)
= -\sum_{k=1}^{\nb}\mathbf{\hat{g}}^\transpose\ \lever_k(\pos),
\end{align}
where $\mathbf{\hat{g}}$ is the gravity acceleration and $\lever(\pos):\confManif\to\R^\nq$ is the $k$-th body's center of mass.

Similarly to the generalized torques, the total energy of the system (kinetic + potential) is affine to the inertial parameters, i.e., 
\begin{align}\label{eq:energy_regressor}
\begin{bmatrix} \KineticEnergy(\pos, \vel; \inertialParams) & \PotentialEnergy(\pos; \inertialParams) \end{bmatrix} = \begin{bmatrix} \regKineticMatrix(\pos,\vel) & \regPotentialMatrix(\pos) \end{bmatrix}^{\transpose} \inertialParams,
\end{align}
where $\regKineticMatrix : \confManif\times\confTManif\to\R^{10\nb}$ and $\regPotentialMatrix : \confManif\to\R^{10\nb}$ are the kinetic and potential energy regressors, respectively.


\subsection{Analytical Derivatives of Energy Regressors}

Deriving the first expression of~\Cref{eq:kinE} can be computationally expensive as it requires to perform a tensor-vector multiplication.
Instead, we derive the second expression as it exploits the structure of the dynamics and allows us to reuse computations, thus:
\begin{align}\label{eq:kinE_dx}
\begin{bmatrix}
\frac{\partial \KineticEnergy}{\partial\pos} \\ \frac{\partial \KineticEnergy}{\partial\vel}
\end{bmatrix} = \sum_{k=0}^{\nb} \boldsymbol{v}_k^\transpose \textbf{I}_k \begin{bmatrix}
\mathcal{X}_k^{-1}\mathbf{J} \\ \mathcal{X}_k^{-1}\mathbf{\dot{J}}
\end{bmatrix},
\end{align}
where $\mathbf{J}$ is the Jacobian that maps joint velocities to spatial velocities \cite{carpentier2019pinocchio} and $\mathbf{\dot{J}}$ is its time derivative. 
The remaining terms can be easily obtained as:
\begin{align}\label{eq:kinE_dq}
\frac{\partial \PotentialEnergy (\pos; \inertialParams)}{\partial\pos} = \textbf{g} \quad
\frac{\partial \KineticEnergy(\pos, \vel; \inertialParams)}{\partial\inertialParams} = \regKineticMatrix \quad
\frac{\partial \PotentialEnergy(\pos; \inertialParams)}{\partial\inertialParams} = \regPotentialMatrix
\end{align}
with $\mathbf{g}(\pos):\confManif\to\R^\nq$ as the gravity vector, collecting the generalized gravity forces.

In general, identifying dynamic parameters requires processing a large number of data points, often on the order of hundred of thousands. 
This, in turn, necessitates an efficient solution strategy for the resulting optimization problem. 
Below, we describe an approach that exploits both the temporal and parametric structure of the problem through a parameterized Riccati recursion.

%% file: chapters/PDDP.tex
\section{Parametrized Equality Constrained Riccati Recursion}\label{sec:parametrized_riccati}

Efficiently handling the stagewise equality constraints 
$\GeqFunc(\state[k],\inputVar[k];\params\vert\ctrlMeas[k])=\zeroVec$ 
together with the parameter constraints 
$\GeqFuncP(\params)=\zeroVec$ and $\GineqFuncP(\params)\ge\zeroVec$ 
in~\Cref{eq:oe_problem} requires a specialized Riccati solver that exploits both the \emph{temporal and parametric structure} of the~\gls{sysid} problem. 
To leverage its temporal structure, we first analyze the optimality conditions through a Bellman perspective, which leads to an equality-constrained, parameterized Riccati recursion. 
This recursion propagates a coupled state-parameter value function backward in time.

\subsection{Optimality conditions}
By examining the Bellman equation associated with~\Cref{eq:oe_problem} under the above assumptions, specifically,
\begin{gather}\label{eq:bellman_eq}
\begin{aligned}
\valFunc\left(\state;\params\right\vert\ctrlMeas,\obsMeas) =& \min_{\state^\prime,\state,\inputVar,\params}\ell(\state,\inputVar,\params\vert\ctrlMeas,\obsMeas)+\valFunc^\prime(\state^\prime;\params\vert\ctrlMeas,\obsMeas)\\
\quad\text{subject to}\quad \state^\prime &= \dynFunc(\state,\inputVar;\params\vert\ctrlMeas) \hspace{2em}\eqFunc(\state,\inputVar;\params\vert\ctrlMeas) = \zeroVec,
\end{aligned}
\end{gather}
we break the~\gls{sysid} problem (in a multiple-shooting sense) into a sequence of subproblems, where the \textit{prime} superscript denotes the next node.
The~\gls{kkt} point for each subproblem can be efficiently determined using the Newton method {(i.e., $\nabla\mathbf{r}\,\delta\mathbf{r}=-\mathbf{r}$)}, yielding the linear system of equations:
\begin{align}\label{eq:kkt}\nonumber
&\overbrace{\begin{bmatrix}
\lagHess[\state\state]               & \lagHess[\state\inputVar]              & \lagHess[\state\params]          & \dynFunc[\state]^\transpose     & \eqFunc[\state]^\transpose &\\
\lagHess[\state\inputVar]^\transpose & \lagHess[\inputVar\inputVar]           & \lagHess[\inputVar\params]       & \dynFunc[\inputVar]^\transpose  & \eqFunc[\inputVar]^\transpose &\\
\lagHess[\state\params]^\transpose   &  \lagHess[\inputVar\params]^\transpose & \lagHess[\params\params]         &\dynFunc[\params]^\transpose       & \eqFunc[\params]^\transpose &  \valFunc[\state\params]^{\prime\transpose} \\
\dynFunc[\state]                     & \dynFunc[\inputVar]                    &\dynFunc[\params]              &                                 &                             & - \mathbf{I} \\
\eqFunc[\state]                     & \eqFunc[\inputVar]                    & \eqFunc[\params]                &                                 &                             &   \\
                                     &                                        & \valFunc[\state\params]^{\prime} & - \mathbf{I}                    &                             & \valFunc[\state\state]^{\prime} 
\end{bmatrix}}^{{\nabla\mathbf{r}}}
\overbrace{\begin{bmatrix}
\delta \state \\ \delta \inputVar \\ \delta \params \\ \mulpDynNext \\ \mulpEqNext \\ \delta \state^{\prime} 
\end{bmatrix}}^{{\delta\mathbf{r}}} =-
\overbrace{\begin{bmatrix}
\costGrad[\state] \\
\costGrad[\inputVar] \\
\costGrad[\params] + \valFunc[\params]^\prime \\
{\dynFeas} \\
{\eqFeas} \\
 \valFunc[\state]^{\prime}
\end{bmatrix}}^{{\mathbf{r}}}\\
&\text{with:}\\\nonumber
&\hspace{0.5em}\mulpDynNext \coloneqq \mulpDyn + \delta\mulpDyn, \hspace{5.8em}
{\dynFeas \coloneqq \dynFunc(\state, \inputVar) \ominus \state^{\prime}},\\\nonumber
&\hspace{0.5em}\mulpEqNext \coloneqq \mulpEq + \delta\mulpEq, \hspace{5.8em}
{\eqFeas \coloneqq \eqFunc(\state, \inputVar; \params)},\\\nonumber
&\hspace{0.5em}\lagHess[\state\state] \coloneqq \costGrad[\state\state] + \valFunc[\state]^{\prime} \cdot \dynFunc[\state\state], \hspace{2em}
\lagHess[\state\inputVar] \coloneqq \costGrad[\state\inputVar] + \valFunc[\state]^{\prime} \cdot \dynFunc[\state\inputVar],\\\nonumber
&\hspace{0.5em}\lagHess[\state\params] \coloneqq \costGrad[\state\params] + \valFunc[\state]^{\prime} \cdot \dynFunc[\state\params], \hspace{2em}
\lagHess[\inputVar\inputVar] \coloneqq \costGrad[\inputVar\inputVar] + \valFunc[\state]^{\prime} \cdot \dynFunc[\inputVar\inputVar],\\\nonumber
&\hspace{0.5em}\lagHess[\inputVar\params] \coloneqq \costGrad[\inputVar\params] + \valFunc[\state]^{\prime} \cdot \dynFunc[\inputVar\params], \hspace{1.5em}
\lagHess[\params\params] \coloneqq \costGrad[\params\params] + \valFunc[\params\params]^{\prime} + \valFunc[\state]^{\prime} \cdot \dynFunc[\params\params],
\end{align}
where $\mathbf{r}$ represents the residual vector containing the gradient of the Lagrangian of~\Cref{eq:bellman_eq}.
Moreover, $\costGrad[\pvar]$, $\dynFunc[\pvar]$, $\eqFunc[\pvar]$ are the first derivative of the cost, the system dynamics and equality constraint with respect to $\pvar$, with $\pvar$ a hypothetical decision variable that represents $\state$, $\inputVar$ or $\params$; $\costGrad[\pvar\pvar]$, $\dynFunc[\pvar\pvar]$, $\eqFunc[\pvar\pvar]$ are the second derivatives; $\valFunc[\pvar]^{\prime}$, {$\valFunc[\pvar\pvar]^{\prime}$ are} the gradient and Hessian of the value function; $\dynFeas$ describe the infeasibility in the integrator; $\mulpDyn$ is the Lagrange multiplier associated to the integrator; $\eqFeas$ describe the infeasibility of the constraint (including implicit dynamics); $\mulpEq$ is the Lagrange multiplier associated to the constraints.
Note that $\delta\state$, $\delta\inputVar$, $\delta\params$, $\delta\state^\prime$, $\delta\mulpDyn$, and $\delta\mulpEq$ provides the search direction computed for the primal and dual variables, respectively.

Similar to optimal control~\cite[Section 2.2]{mastalli22auro}, we observe the presence of the Markovian structure, leading to ${\mulpDynNext = \valFunc[\state]^\prime + \begin{bmatrix} \valFunc[\state\state]^\prime & \valFunc[\state\params]^\prime\end{bmatrix}\begin{bmatrix}\delta\state^\prime \\ \delta\params\end{bmatrix}}$.
This relationship allows us to condense this linear system of equations and describe this problem as
\begin{align}\label{eq:condensed_kkt}
&\Delta\valFunc = \nonumber\\ &\min_{\delta\inputVar,\delta\params} 
\begin{bmatrix}
\delta\state\\
\delta\inputVar \\
\delta\params    
\end{bmatrix}^\transpose
\begin{bmatrix}
\qualFunc[\state\state] & \qualFunc[\state\inputVar] & \qualFunc[\state\params] \\
\qualFunc[\state\inputVar]^\transpose & \qualFunc[\inputVar\inputVar] & \qualFunc[\inputVar\params] \\
\qualFunc[\state\params]^\transpose & \qualFunc[\inputVar\params]^\transpose & \qualFunc[\params\params]
\end{bmatrix}
\begin{bmatrix}
\delta\state\\
\delta\inputVar \\
\delta\params    
\end{bmatrix}
+\begin{bmatrix}
\delta\state\\
\delta\inputVar \\
\delta\params    
\end{bmatrix}^\transpose
\begin{bmatrix}
\qualFunc[\state]\\
\qualFunc[\inputVar] \\
\qualFunc[\params]   
\end{bmatrix},\nonumber\\
&\quad\text{subject to}\quad \begin{bmatrix}\eqFunc[\state]&\eqFunc[\inputVar]&\eqFunc[\params]\end{bmatrix} \begin{bmatrix}\delta\state\\ \delta\inputVar\\ \delta\params\end{bmatrix} + \eqFeas = \zeroVec,
\end{align}
where the $\qualFunc$'s terms represent the local approximation of the unconstrained \textit{action-value function} whose expressions are:
\begin{align}\nonumber
&\small\qualFunc[\state\state] = \lagHess[\state\state] + \dynFunc[\state]^\transpose \valFunc[\state\state]^\prime \dynFunc[\state],
\hspace{2.1em}
\qualFunc[\inputVar\params] = \lagHess[\inputVar\params] + \dynFunc[\inputVar]^\transpose (\valFunc[\state\params]^\prime + \valFunc[\state\state]^\prime \dynFunc[\params]), \\\nonumber
&\small\qualFunc[\state\inputVar] = \lagHess[\state\inputVar] + \dynFunc[\state]^\transpose \valFunc[\state\state]^\prime \dynFunc[\inputVar], 
\hspace{1.7em}
\qualFunc[\state\params] = \lagHess[\state\params] + \dynFunc[\state]^\transpose (\valFunc[\state\params]^\prime + \valFunc[\state\state]^\prime \dynFunc[\params]), \\\nonumber
&\small\qualFunc[\inputVar\inputVar] = \lagHess[\inputVar\inputVar] + \dynFunc[\inputVar]^\transpose \valFunc[\state\state]^\prime \dynFunc[\inputVar], \hspace{1.2em}
\qualFunc[\params\params] = \lagHess[\params\params] + \dynFunc[\params]^\transpose (2\valFunc[\state\params]^\prime + \valFunc[\state\state]^\prime \dynFunc[\params]), \\\nonumber
&\small\qualFunc[\state] = \costGrad[\state] + \dynFunc[\state]^\transpose {\valFunc[\state]^+},
\hspace{4.9em}
\qualFunc[\params] = \costGrad[\params] + \valFunc[\params]^+ + \dynFunc[\params]^\transpose {\valFunc[\state]^+},\\
&\small\qualFunc[\inputVar] = \costGrad[\inputVar] + \dynFunc[\inputVar]^\transpose {\valFunc[\state]^+}
\end{align}
with ${\valFunc[\state]^+ \coloneqq \valFunc[\state]^\prime + \valFunc[\state\state]^\prime \dynFeas}$ and ${\valFunc[\params]^+ \coloneqq \valFunc[\params]^\prime + \valFunc[\state\params]^\prime \dynFeas}$ as the gradients of the value function after the deflection produced by the dynamics infeasibility $\dynFeas$ (see~\cite{mastalli-icra20}).
Computing $\delta\inputVar$ and $\delta\params$ can be done by building an estimation policy as described next. 

\subsection{Constrained policy and value function}
Contrary to traditional~\gls{ddp}, finding a solution to~\Cref{eq:condensed_kkt} requires to solve a constrained \gls{qp}, which has the following optimality conditions:
\begin{align}\label{eq:optimallity_u}
\begin{bmatrix}
    \qualFunc[\inputVar\inputVar] &\eqFunc[\inputVar]^\transpose \\ 
    \eqFunc[\inputVar] &
\end{bmatrix}
\begin{bmatrix}
    \delta\inputVar \\ 
    \mulpEqNext
\end{bmatrix} = -
\begin{bmatrix}
    \qualFunc[\inputVar] + \qualFunc[\inputVar\state]\delta\state + \qualFunc[\inputVar\params]\delta\params\\
    \eqFeas + \eqFeas[\state]\delta\state + \eqFunc[\params]\delta\params
\end{bmatrix}.
\end{align}

\Cref{eq:optimallity_u} can be solved using the \textit{Schur-complement} factorization; however, this approach increases the algorithm complexity of the Riccati recursion. This increment is related to the number of equality constraints. 
Instead, as proposed in work \cite{mastalli-invdynmpc} we follow a nullspace factorization approach because it does not increase the algorithm complexity needed to handle the equality constraint.

\begin{figure*}[t]\centering
    \href{\video&t=26}{\includegraphics[width=0.24\linewidth,valign=t]{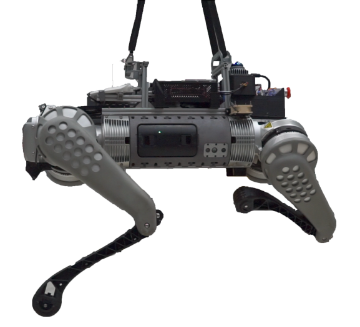}
    \includegraphics[width=0.24\linewidth,valign=t]{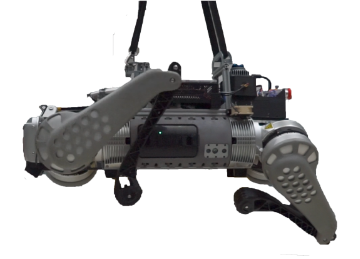}
    \includegraphics[width=0.24\linewidth,valign=t]{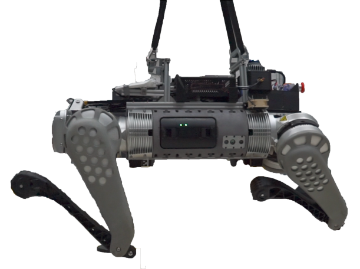}
    \includegraphics[width=0.24\linewidth,valign=t]{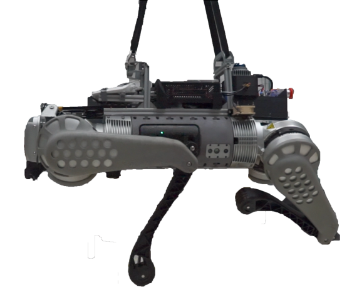}
        \includegraphics[width=0.24\linewidth,valign=b]{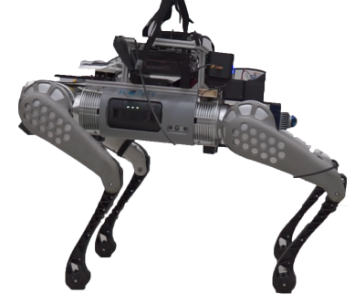}
        \includegraphics[width=0.24\linewidth,valign=b]{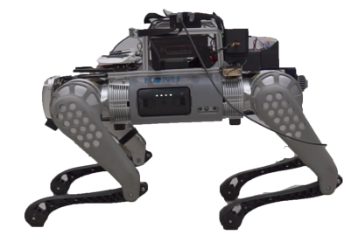}
        \includegraphics[width=0.24\linewidth,valign=b]{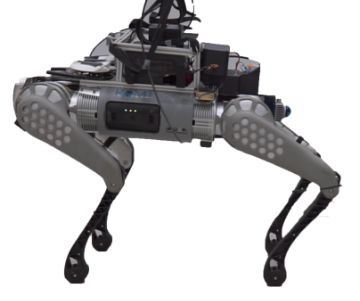}
        \includegraphics[width=0.24\linewidth,valign=b]{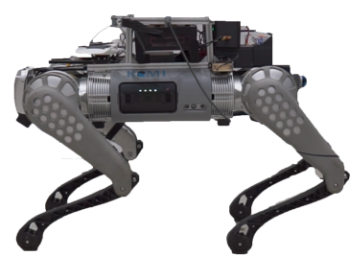}}
    \caption{ Optimal excitation motions executed on the Unitree B1 quadruped. Top: excitation of a single leg while the rest of the robot remains fixed. Bottom: excitation of the full floating-base system. These trajectories are generated by the proposed framework to maximize parameter observability while respecting actuation limits and kinematic constraints.}
    \label{fig:snapshots_excitation}
\end{figure*}

\subsubsection{Nullspace factorization}
We can decompose the search direction of $\delta\inputVar$ as:
\begin{align}\label{eq:control_decomposition}
\delta\inputVar = \rankSpace \delta\inputVar[\rank] + \NullSpace \delta\inputVar[\Null],
\end{align}
where $\NullSpace \in \R^{\ntau \times n_z}$ is the nullspace basis of $\eqFunc[\inputVar]$ and $\rankSpace \in \R^{\ntau \times n_y}$ is its orthogonal matrix.
Then, by substituting this parametrization into~\Cref{eq:optimallity_u} and observing that $\eqFunc[\inputVar] \mathbf{Z} = \zeroVec$, we obtain:
\begin{align}\label{eq:nullspace}
\begin{bmatrix}
    \NullSpace^\transpose \qualFunc[\inputVar\inputVar] \NullSpace &\rankSpace^{\transpose} \qualFunc[\inputVar\inputVar]\rankSpace \\ 
    & \eqFunc[\inputVar]\rankSpace
\end{bmatrix}&
\begin{bmatrix}
    \delta\inputVar[\Null] \\ 
    \delta\inputVar[\rank]
\end{bmatrix} =\\\nonumber &-
\begin{bmatrix}
    \NullSpace^{\transpose}(\qualFunc[\inputVar] + \qualFunc[\inputVar\state]\delta\state + \qualFunc[\inputVar\params]\delta\params)\\
    \eqFeas + \eqFeas[\state]\delta\state + \eqFunc[\params]\delta\params
\end{bmatrix},
\end{align}
which allows us to compute the control policy without explicitly computing $\mulpEqNext$.
This yields to the following \emph{estimation policy}:
\begin{align}\label{eq:control_PolicyNull}
& \quad \delta\inputVar = - \ffPolicyFinal - \fbPolicyFinal \delta\state - \pPolicyFinal \delta\params\\\nonumber
\noindent  \text{with:}\\\nonumber
\ffPolicyFinal &= \NullSpace \ffPolicy + \qualFuncConst[\Null\Null]\boldsymbol{\Psi}_{n} \eqFeas \quad \text{(feed forward),}\\\nonumber
\fbPolicyFinal &= \NullSpace \fbPolicy + \qualFuncConst[\Null\Null]\boldsymbol{\Psi}_{n} \eqFunc[\state] \quad \text{(state feedback gain),}\\\nonumber
\pPolicyFinal &= \NullSpace \pPolicy + \qualFuncConst[\Null\Null]\boldsymbol{\Psi}_{n} \eqFunc[\params] \quad \text{(parameters feedback gain),}\\\nonumber
\end{align}
where $\ffPolicy = \qualFunc[\Null\Null]^{-1}\qualFunc[\Null]$, $\fbPolicy = \qualFunc[\Null\Null]^{-1}\qualFunc[\Null\state]$ and  $\pPolicy = \qualFunc[\Null\Null]^{-1}\qualFunc[\Null\params]$ are the feed-forward and feedback gain associated to the nullspace of the equality constraint, $\boldsymbol{\Psi}_{n} = \rankSpace(\eqFunc[\inputVar]\rankSpace)^{-1}$ and $\qualFuncConst[\Null\Null] = \mathbf{I} - \NullSpace\qualFunc[\Null\Null]^{-1}\qualFunc[\Null\inputVar]$ are terms that project the constraint into both spaces:
nullspace and range. Finally, $\qualFunc[\Null] = \NullSpace^\transpose\qualFunc[\inputVar]$, $\qualFunc[\Null\inputVar] = \NullSpace^\transpose\qualFunc[\inputVar\inputVar]$, 
$\qualFunc[\Null\state] = \NullSpace^\transpose\qualFunc[\inputVar\state]$, 
$\qualFunc[\Null\params] = \NullSpace^\transpose\qualFunc[\inputVar\params]$, are the local approximations of the action-value function in the nullspace.
Note that the stagewise nullspace decompositions can be done in parallel without increasing the algorithmic complexity of the Riccati recursion as explained in~\cite{mastalli-invdynmpc}.

When plugging the changes in the estimation policy into~\Cref{eq:condensed_kkt}, we obtain a quadratic approximation of the value function as follows
\begin{align}\nonumber
\delta\valFunc(\delta\state; \delta\params\vert\ctrlMeas,\obsMeas)& \simeq \Delta\valFunc[1] + \frac{\Delta\valFunc[2]}{2}\\
&\hspace{-4em} + \frac{1}{2}
\begin{bmatrix}
\delta\state\\
\delta\params
\end{bmatrix}^\transpose
\begin{bmatrix}
\valFunc[\state\state] & \valFunc[\state\params] \\
\valFunc[\state\params]^\transpose & \valFunc[\params\params]
\end{bmatrix}
\begin{bmatrix}
\delta\state\\
\delta\params
\end{bmatrix}
+ \begin{bmatrix}
\valFunc[\state]\\
\valFunc[\params]
\end{bmatrix}^\transpose
\begin{bmatrix}
\delta\state\\
\delta\params
\end{bmatrix},
\end{align}
where\vspace{-0.8em}
\begin{align}\nonumber
& \Delta\valFunc[1] = - \ffPolicyFinal^\transpose \qualFunc[\inputVar], \quad\quad\quad\,\, \Delta\valFunc[2] = \ffPolicyFinal^\transpose\qualFunc[\inputVar\inputVar]\ffPolicyFinal, \\\nonumber
&\valFunc[\state] = \qualFunc[\state] + \fbPolicyFinal^\transpose (\qualFunc[\inputVar\inputVar]\ffPolicyFinal-\qualFunc[\inputVar]) - \qualFunc[\inputVar\state]^\transpose\ffPolicyFinal,  \\\nonumber
&\valFunc[\params] = \qualFunc[\params] - \qualFunc[\params\inputVar]\ffPolicyFinal + \pPolicyFinal[][\transpose](\qualFunc[\inputVar\inputVar]\ffPolicyFinal-\qualFunc[\inputVar]), \\\nonumber
&\valFunc[\state\state] = \qualFunc[\state\state] + \fbPolicyFinal^\transpose (\fbPolicyFinal^{\transpose}\qualFunc[\inputVar\inputVar]-2\qualFunc[\inputVar\state]^{\transpose}), \\\nonumber
&\valFunc[\params\params] = \qualFunc[\params\params] + \pPolicyFinal[][\transpose](\qualFunc[\inputVar\inputVar]\pPolicyFinal-2\qualFunc[\inputVar\params]), \\\nonumber
&\valFunc[\state\params] = \qualFunc[\state\params] + \fbPolicy^{\transpose}(\qualFunc[\inputVar\inputVar]\pPolicyFinal-\qualFunc[\inputVar\params]) - \qualFunc[\state\inputVar]\pPolicyFinal. \\\nonumber
\end{align}
With this value function, we propagate backward in time until reaching the arrival (initial) node.
We next describe the computations performed at the arrival node, which, compared to optimal control algorithms, is unknown.

\subsection{Arrival node, parameter update and parameter constraints}
The parameterized Riccati recursion yields a quadratic approximation of the value function at the arrival node.
We first solve for the feedforward parameter perturbation by considering only the parameter block of this approximation, which leads to the constrained~\gls{qp}
\begin{align}\label{eq:constrained_qp}\nonumber
\min_{\delta\params} & \quad \frac{1}{2}\delta\params^\transpose \valFunc[\params\params]^\bullet \delta\params
 + \valFunc[\params]^{\bullet\,\transpose} \delta\params \\
\textrm{subject to} & \quad \GeqFuncP\delta\params + \eqFeasP = \zeroVec, \nonumber\\
& \quad \GineqFuncP\delta\params + \ineqFeasP \geq \zeroVec ,
\end{align}
where the superscript $\bullet$ denotes arrival-node quantities. 
The matrices $\GeqFuncP$ and $\GineqFuncP$ collect the Jacobians of the parameter-consistency constraints, while $\eqFeasP$ and $\ineqFeasP$ measure their current infeasibility.
Solving~\Cref{eq:constrained_qp} yields the feedforward term $\delta\params[\text{ff}]$, which we obtain with the warm-startable solver proposed in~\cite{jrojas-odyn25}.

To incorporate the coupling with the arrival state, we compute the feedback gain $\fbPolicyParam$ by reusing the same factorization.
Specifically, the derivatives of the~\gls{kkt} system with respect to $\delta\state[0]$ require solving linear systems of the form $\valFunc[\params\params]^\bullet \mathbf{v} = -\,\valFunc[\state\params]^{\bullet\,\transpose}$ (with the constraint Jacobians appended as in the~\gls{qp}). 
Because the factorization of $\valFunc[\params\params]^\bullet$ and the constraint matrices is already available from~\Cref{eq:constrained_qp}, we obtain the feedback gain at negligible additional cost and express the parameter policy as
\begin{equation}
\delta\params = \delta\params[\text{ff}] + \fbPolicyParam\,\delta\state[0].
\end{equation}

Once the parameter policy is available, the arrival-state perturbation is recovered from the linear system
\begin{equation}
\valFunc[\state\state]^\bullet\,\delta\state[0] = -\left(\valFunc[\state]^\bullet + \valFunc[\state\params]^\bullet\,\delta\params\right),
\end{equation}

%% file: chapters/results.tex
\section{Results}\label{sec:results}
\begin{figure}[b]\centering
    \href{\video&t=204}{\includegraphics[width = 1.\linewidth]{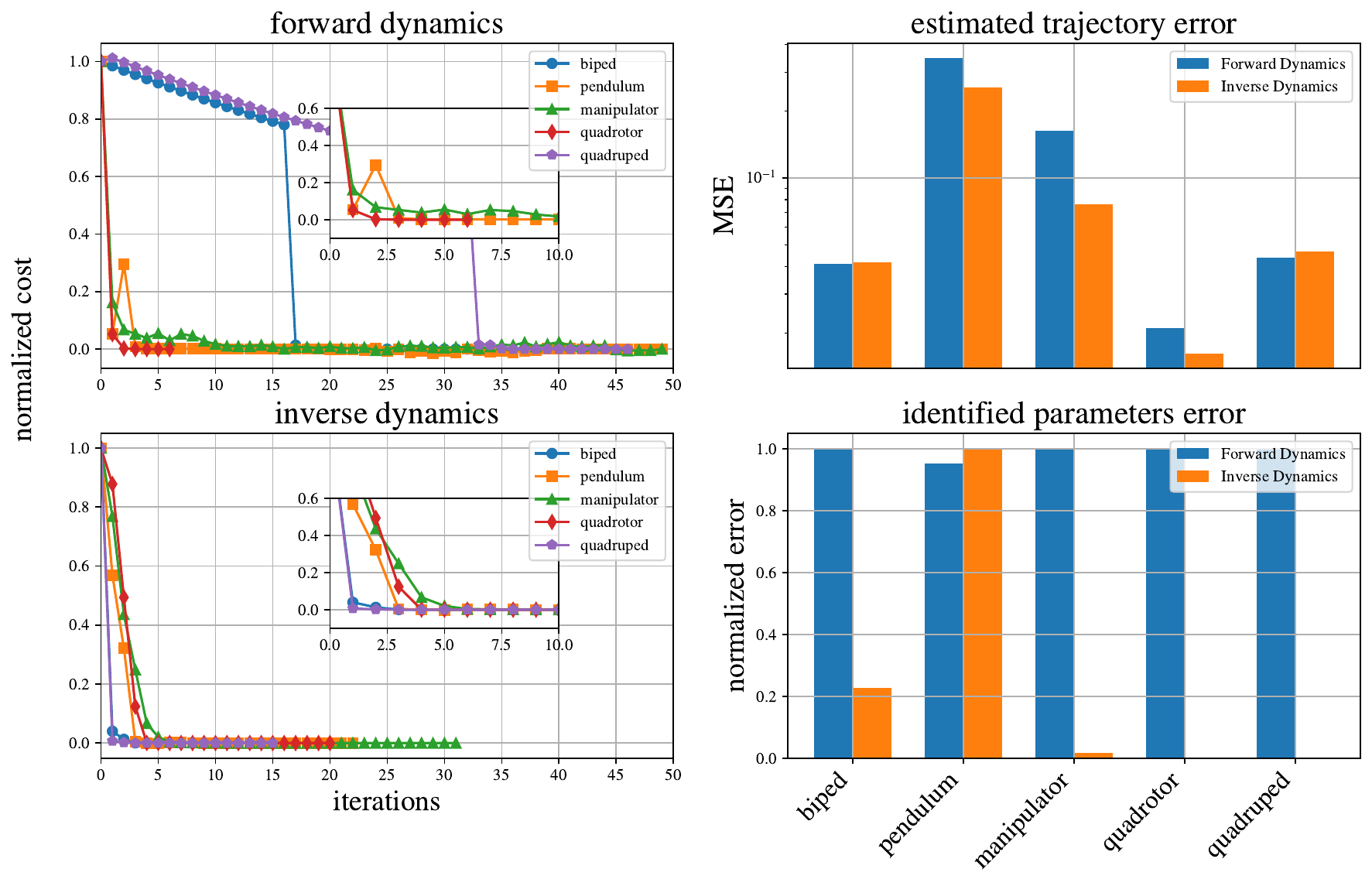}}
    \caption{Cost, estimated trajectory errors, and inertial parameters errors for inverse and forward dynamics formulations. 
    The inverse dynamics formulation shows faster convergence and lower estimation error of both the state trajectory and inertial parameters when compared with the forward dynamics formulation.}
    \label{fig:inv_vs_fwd}
\end{figure}


\begin{figure}[b]\centering
    \href{\video&t=218}{\includegraphics[width = 1.\linewidth]{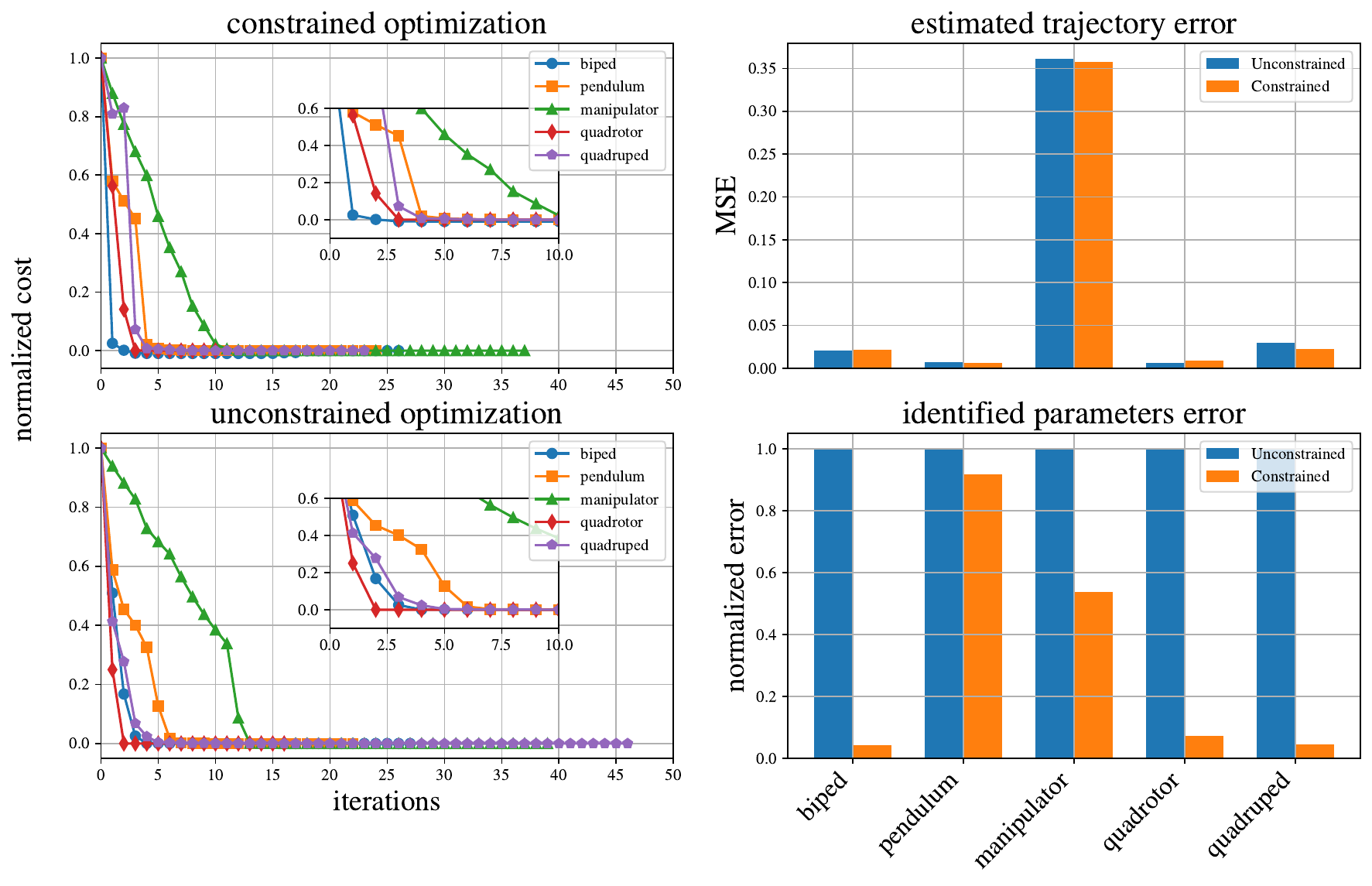}}
    \caption{Evolution of the cost, trajectory estimation error, and inertial parameter identification error for both constrained and unconstrained~\gls{sysid} formulations. 
    The addition of parameter constraints maintains similar convergence behavior and trajectory accuracy compared to the unconstrained case, while notably enhancing the precision of the identified inertial parameters. 
    The improvement is especially pronounced in platforms exhibiting discrete symmetries, such as bipeds and quadrupeds.}
\label{fig:constrained_params}
\end{figure}

\begin{figure*}[t]\centering
    \href{\video&t=346}{\includegraphics[width=0.24\linewidth,valign=t]{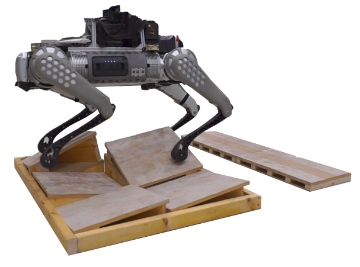}
    \includegraphics[width=0.24\linewidth,valign=t]{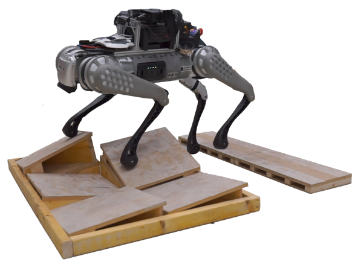}
    \includegraphics[width=0.24\linewidth,valign=t]{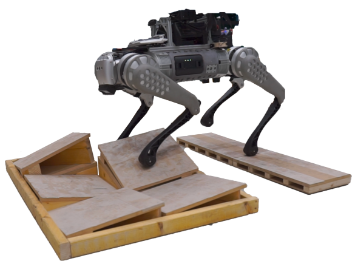}
    \includegraphics[width=0.24\linewidth,valign=t]{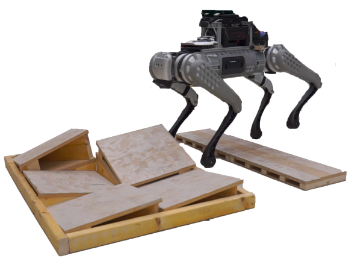}
    \includegraphics[width=0.22\linewidth,valign=t]{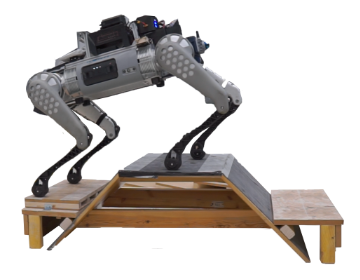}
    \includegraphics[width=0.21\linewidth,valign=t]{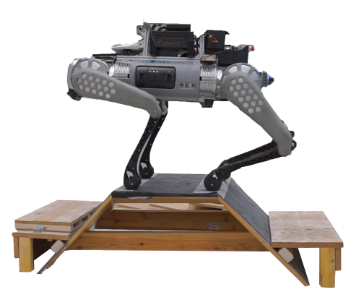}
    \includegraphics[width=0.26\linewidth,valign=t]{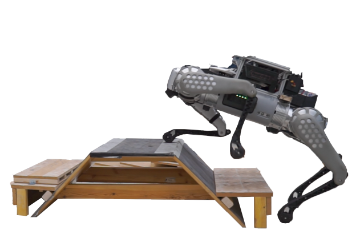}
    \includegraphics[width=0.28\linewidth,valign=t]{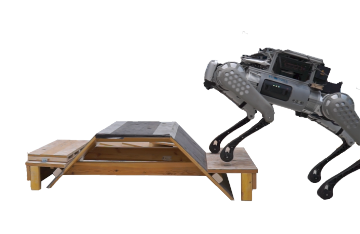}
    \includegraphics[width=0.27\linewidth,valign=b]{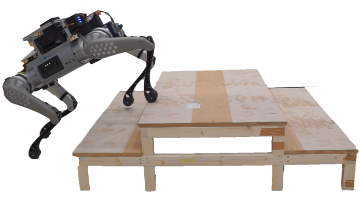}
    \includegraphics[width=0.24\linewidth,valign=b]{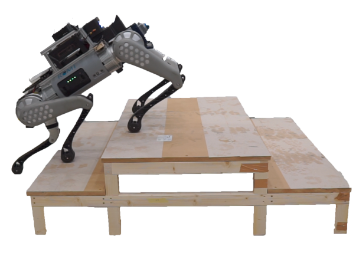}
    \includegraphics[width=0.23\linewidth,valign=b]{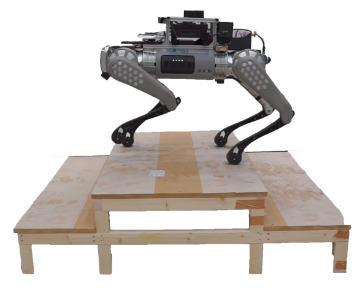}
    \includegraphics[width=0.24\linewidth,valign=b]{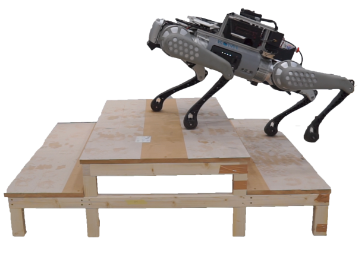}}
    \caption{Validation of the identified model on the Unitree B1 quadruped. Top: traversal of rough terrain. Middle: stair climbing. Bottom: climbing bigger stairs. The motions are executed using the model identified from optimal excitation trajectories, demonstrating transfer and generalization to locomotion tasks beyond the identification motions. }
    \label{fig:snapshots_results}
\end{figure*}

In this section, we analyze our~\gls{sysid} framework through a series of simulation and hardware experiments. 
The experiments are designed to validate the framework's ability to simultaneously estimate the state trajectory and physical parameters of robotic systems under realistic conditions, including noise, contact dynamics, and closed-loop constraints. 
We compare the performance of the inverse dynamics formulation against the forward dynamics formulation, analyze the impact of incorporating parameter constraints, and demonstrate the benefits of energy-based observations for improving parameter identifiability. 
Finally, we validate the framework on hardware platforms, including the Unitree B1 quadruped and Z1 manipulator, showcasing its practical applicability to real-world robotic systems.

\subsection{Problem Specifications and Setup}\label{sec:setup}
We evaluated the our~\gls{sysid} framework in simulation using dynamically-consistent, noisy measurements generated from simulated trajectories. 
Reference trajectories were synthesized by solving the optimal-excitation program described in Appendix~\ref{sec:opt_traj} (see also \Cref{eq:opt_traj}), and then propagated with the multibody dynamics in ~\Cref{eq:dyn,eq:kkt_acc,eq:kkt_vel_reset}. 
This yielded ground-truth state and parameter-consistent signals against which we tested system identification as shown in~\Cref{fig:snapshots_excitation}.

For each experiment, we provided the Bayesian estimator with a standard set of proprioceptive measurements:
(i) simulated joint encoders $(\posMeas,\velMeas)$,
(ii) commanded joint efforts $\ctrlMeas$ used by the actuation model,
and (iii) a torso IMU (linear acceleration $\accMeas$, body angular rate $\angVelMeas$, and orientation) in floating-base robots.
Measurements were corrupted by zero-mean Gaussian noise with covariances chosen to mimic typical datasheets. 
Formally, at each node $k$, the observation model was $\obsMeas[k] = \obsFunc(\state[k];\params)$ with $\epsilon_{k}\!\sim\!\mathcal{N}(0,\Sigma_{k})$ and the system parameters were initialized with a $70\%$ relative uncertainty.

\subsection{Forward vs Inverse Dynamics Formulations}
We compared forward dynamics and inverse dynamics formulations across five robotic systems, a double pendulum, manipulator, quadrotor, biped, and quadruped.
We provided the Bayesian estimator with the same measurements, noise, priors, horizon, without enforcing additional constraints.

\Cref{fig:inv_vs_fwd} summarizes normalized cost decay over iterations, state-trajectory error and parameter error.  
Across all systems, inverse dynamics converges in fewer iterations than forward dynamics, with the gap most pronounced for contact-rich cases (biped, quadruped).
This is because inverse dynamics’ explicit handling of constraint consistency prevented the estimator from ``bending'' parameters to explain the mismatch between the dynamics and observations.

The trajectory error is often smaller under inverse dynamics, as forward dynamics more often compensates the mismatch between dynamics and observations by injecting process noise in the trajectory, resulting in higher costs and parameter errors.
This is because inverse dynamics decouples numerical integration from the system dynamics and preserves the affine dependence of the constraints on the inertial parameters, thereby improving numerical conditioning.
This also enforces contact consistency at each step, avoiding bias in both state and parameters. 

\subsection{Benefits of Parameter Constraints}\label{sec:benefits_param_constraints}
We compared our inverse-dynamics~\gls{sysid} approach with a parameter-constrained variant that enforces the parameter equalities/inequalities in ~\Cref{eq:oe_problem:PEQ,eq:oe_problem:PINEQ}.
We chose to enforce left-right inertial symmetries on mirrored limbs and total mass constraints.
As reported in~\Cref{fig:constrained_params}, the constrained and unconstrained~\gls{sysid} formulations show nearly identical convergence, showing stable numerical capabilities of our~\gls{sysid} optimizer.
However, the constrained formulation consistently yields smaller inertial‐parameter errors. 
The advantage is most evident on symmetric platforms (biped, quadruped), where including the concept of discrete symmetries removes redundant directions~\cite{ordonez2024morphological}.

\subsection{Effect of Energy-Based Observations}\label{sec:ablation_energy}
We evaluated the impact of energy-based observation by averaging $100$ random initializations for each system. 
\Cref{tab:energy_obs} shows that including energy-based observation changes the optimization landscape.
The added residual introduces a stiffer trade-off between model consistency and measurement fit, which increases the number of iterations by roughly $25$--$75\%$.
This raises the final cost by about one order of magnitude, and slightly increases the trajectory error (because the estimator can no longer accommodate modeling gaps via small energy mismatches).

The main benefit remains parameter identifiability. 
Across the manipulator, biped, and quadruped, we observed a noticeable reduction in friction errors (up to $\approx50\%$) and a mild yet consistent improvement in inertial parameters. 
The pendulum is already well conditioned, so the effect on trajectories is negligible, but the energy-based observations still reduced friction uncertainty by half. 
This means that sligh increase in iterations is acceptable.

\begin{table}[t]\centering
\caption{Effect of energy-based observations on convergence (100 random initial guesses). Entries are means over trials.}
\label{tab:energy_obs}
\setlength{\tabcolsep}{4pt} 
\ra{1.1}
\begin{tabular}{@{} l *{4}{@{\hspace{1.\tabcolsep}} cc} @{}}
& \multicolumn{2}{c}{\textit{Pendulum}} & \multicolumn{2}{c}{\textit{Manipulator}} & \multicolumn{2}{c}{\textit{Biped}} & \multicolumn{2}{c}{\textit{Quadruped}}\\
& w/o & with & w/o & with & w/o & with & w/o & with \\
\hline
\textbf{Iterations}                & 20.4 & 31.2 & 7.3 & 11.4 & 15.6 & 24.1 & 9.2 & 15.5 \\
\textbf{Cost $[\cdot 10^{-1}]$}    & 0.09 & 1.62 & 0.29 & 3.28 & 0.47 & 4.72 & 0.41 & 3.67 \\
\textbf{Traj. Err [$\ell_\infty$]} & 0.41 & 0.46 & 2.58 & 2.83 & 5.88 & 6.34 & 4.76 & 5.08 \\
\textbf{Iner. Err [$\ell_2$]}   & 0.022 & 0.018 & 0.56 & 0.46 & 1.86 & 1.62 & 1.47 & 1.28 \\
\textbf{Fric. Err [$\ell_2$]}   & 0.031 & \textbf{0.016} & 0.75 & \textbf{0.40} & 2.18 & \textbf{1.08} & 1.82 & \textbf{0.91} \\
\midrule
\end{tabular}
\vspace{-0.6em}
\end{table}

\subsection{Comparison with Frequentist Identification Methods}
\label{sec:ablation_frequentist}
To assess the contribution of energy information and the impact of our constrained Bayesian formulation, we compared three identification strategies: (i) a classical inverse-dynamics regressor (ID), (ii) an energy-based regressor using work--energy relations, and (iii) our proposed method, which combines inverse dynamics with energy-based observations inside a constrained Bayesian estimation framework.
All three methods used the same rigid-body model, excitation trajectories, and friction parameterization, and were evaluated under identical noise conditions. 
Importantly, to ensure physical plausibility, both regression baselines were implemented using exponential--eigenvalue inertial parametrization, as in our Bayesian approach.

For the ID-regressor baseline, generalized torques were expressed as a linear function of inertial and actuation parameters, which were estimated by minimizing a least-squares discrepancy between measured and predicted torques.
Since this regressor depends on joint accelerations, accelerations were obtained via numerical differentiation of encoder measurements and filtered using a Savitzky–Golay filter, with the window length tuned empirically. 
For the energy-based regressor, work–energy consistency was enforced over each time interval by matching the measured input work (integrated actuation power) to the model-predicted change in mechanical energy plus dissipated frictional energy.
This formulation provided an identification signal that was less sensitive to acceleration estimation and instead aggregated information across joints.
As shown in Table~\ref{tab:sysID-frequentist}, our approach consistently achieved the lowest identification error across all noise levels.
These results demonstrate that combining inverse-dynamics consistency with energy-based observations within a constrained Bayesian framework improves robustness and mitigates inertial–friction coupling while preserving physical consistency.

\begin{table}[t]
\centering
\caption{Identification errors under increasing Gaussian noise.
Both regression baselines use the EE inertial parametrization to guarantee physical consistency.
ID uses numerically differentiated accelerations filtered with a Savitzky--Golay window tuned to the best of our capabilities.
Energy uses work--energy consistency over each interval.}
\label{tab:sysID-frequentist}
\begin{tabular}{@{}c @{\hspace{1.2\tabcolsep}} ccc@{}}
\toprule
Gaussian noise $\sigma$ & ID (tuned SG) & Energy & Ours \\
\midrule
$0.0$   & $3.5\%$  & $6.5\%$  & $\mathbf{2.0\%}$ \\
$0.005$ & $9.5\%$  & $8.0\%$  & $\mathbf{2.3\%}$ \\
$0.01$  & $22.0\%$ & $12.5\%$ & $\mathbf{2.8\%}$ \\
$0.1$   & $95.0\%$ & $38.0\%$ & $\mathbf{6.0\%}$ \\
\bottomrule
\end{tabular}
\vspace{-1.2\baselineskip}
\end{table}

\begin{figure}[b]\centering
    \begin{minipage}[t]{0.4\linewidth}
        \centering
        \includegraphics[width=\linewidth]{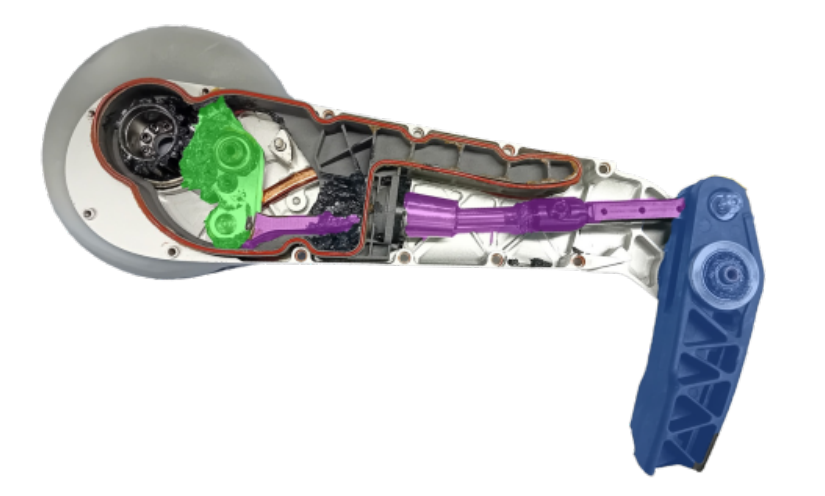}\\
        \subcaption{B1's leg internal mechanism.}
    \end{minipage}
    \begin{minipage}[t]{0.4\linewidth}
        \centering
        \includegraphics[width=\linewidth]{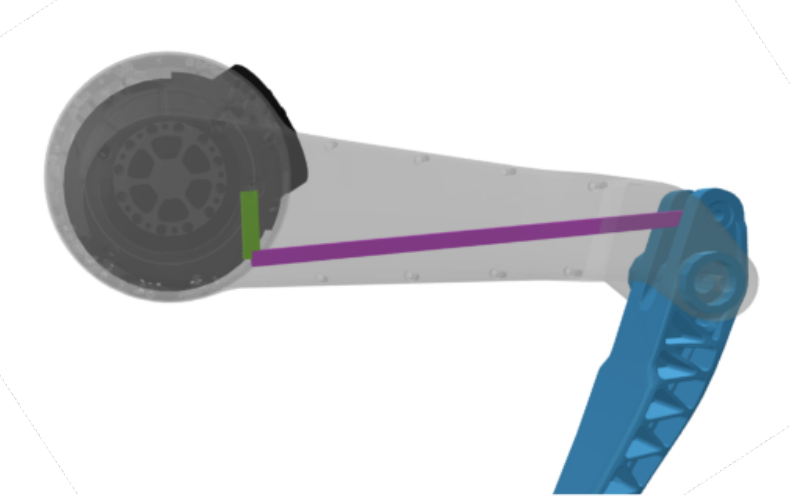}\\
        \subcaption{Model of the kinematic loop.}
    \end{minipage}
    \begin{minipage}[t]{0.48\linewidth}
        \centering
        \includegraphics[width=\linewidth]{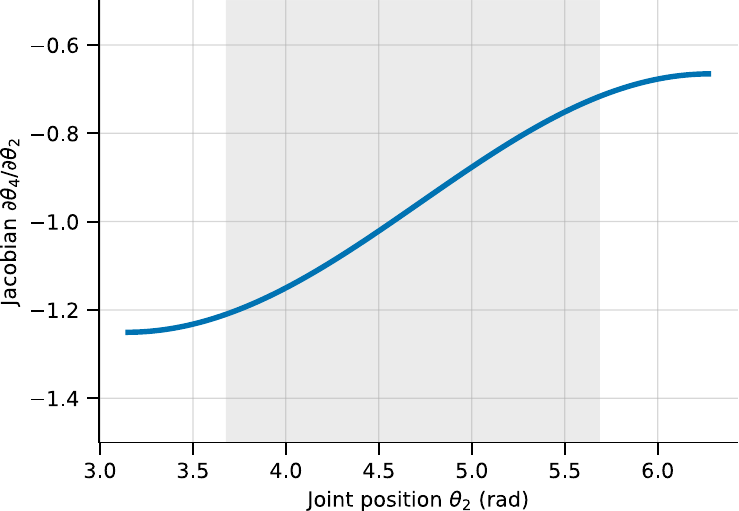}\\
        \subcaption{Evolution of the kinematic Jacobian.}
    \end{minipage}
    \begin{minipage}[t]{0.48\linewidth}
        \centering
        \includegraphics[width=\linewidth]{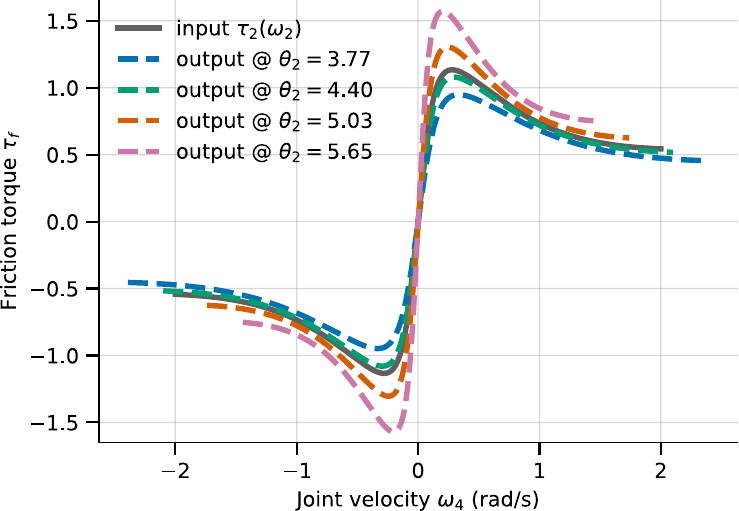}\\
        \subcaption{Motor friction reflected to the output.}
    \end{minipage}
    \vspace{0.5em}
    \caption{
        Visualization of the four-bar mechanism with friction effects: 
        (a) Colored internal mechanism of B1's leg, highlighting the four-bar mechanism; 
        (b) Our model of the four-bar mechanism;
        (c) Evolution of the kinematic Jacobian along the joint position, the highlighted area shows B1's operating range; 
        (d) Joint friction model evolution.
        A hypothetical friction originated in the motor's gearbox is reflected nonlinearly in the output joint following the kinematic Jacobian.
    }
    \label{fig:fourbar_jacobian_friction}
\end{figure}

\subsection{Validation on B1 and Z1 robots}\label{sec:b1z1}
We assessed the \emph{quality} and \emph{generality} of the identified model on hardware with a Unitree B1 quadruped carrying a Unitree Z1 arm. 
We collected measurements from the joint encoders $(\posMeas,\velMeas)$ on legs and arm, torso IMU, and applied torques $\ctrlMeas$. 
Additionally, we enforced bilateral foot contacts, considered the B1 four-bar mechanisms and their joint-friction, and introduced energy observations (\Cref{sec:dyn_inertia,sec:energy}) to improve identifiability without force/torque sensors.

\begin{figure}[t]\centering
    \href{\video&t=158}{\includegraphics[width=1.0\linewidth]{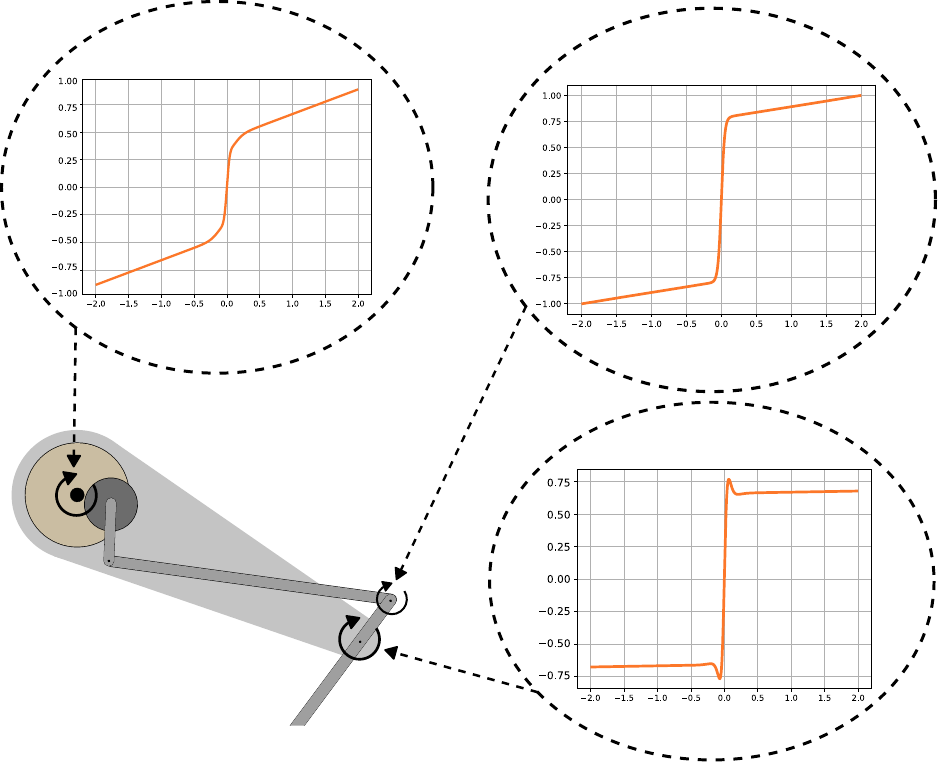}}
    \caption{
        Identified friction characteristics for the KFE joint in the Unitree B1 leg.
        The plots show the normalized estimated joint friction torque as a function of joint velocity for the actuated and passive (non-actuated) joints, present in the four-bar mechanism.
        Each joint exhibits distinct friction profiles, which are explained by a combination of static, viscous, and Stribeck friction effects.
    }
    \label{fig:identified_friction}
\end{figure}

\subsubsection{Mechanism and friction modeling} 
 The B1 leg contains a four-bar transmission that couples joint motions and reflects motor friction to the output. 
~\Cref{fig:fourbar_jacobian_friction} summarizes the hardware and its model: the real mechanism and its loop representation (panels a--b), the evolution of the kinematic Jacobian along the motion (panel c), and the motor-side friction mapped to the output (panel d). 
These elements were embedded in our estimator and enforced at each node. 
Starting from coarse URDF parameters provided by Unitree, the constrained~\gls{sysid} framework converged reliably to physically consistent inertial and friction parameters. 
Importantly, the identified model includes separate friction characteristics (static, viscous, and Stribeck effects) for the actuated and passive (non-actuated) joints within the four-bar mechanism as depicted in~\Cref{fig:identified_friction}.

\subsubsection{Model error and tracking quality}
\begin{figure}[t]
\begin{flushright}
    \begin{minipage}[t]{0.975\linewidth}
        \subcaption{Nominal model}
        \includegraphics[width=\linewidth]{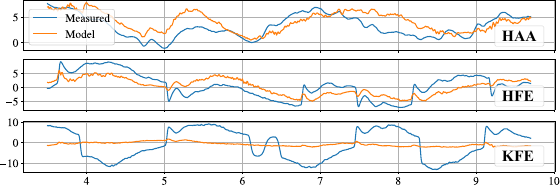}\\
        \vspace{-1.5em}
        \label{fig:nominal_torques}
    \end{minipage}
    \begin{minipage}[t]{1.0\linewidth}
        \subcaption{Identification of inertial and friction parameters in a reduced model.}
        \includegraphics[width=\linewidth]{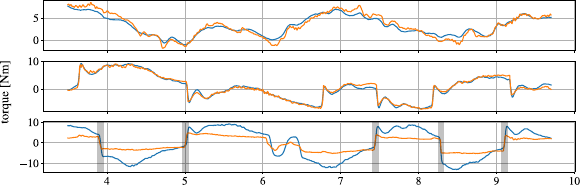}\\
        \vspace{-1.5em}
        \label{fig:identified_model_torques}
    \end{minipage}
    \begin{minipage}[t]{0.975\linewidth}
        \subcaption{Identification of inertial and friction parameters with closed loop kinematics.}
        \includegraphics[width=\linewidth]{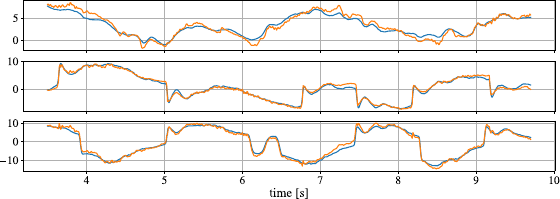}\\
        \label{fig:identified_model_torques_clk}
    \end{minipage}
\end{flushright}
\vspace{-1.5em}
\caption{
    Measured (blue) versus model-predicted (orange) joint torques for a representative trajectory of B1's leg (HAA, HFE, and KFE joints).
    (a) Nominal model obtained from the initial URDF parameters: large torque prediction errors, especially at KFE (approximately one order of magnitude).
    (b) After identifying inertial and friction parameters on a reduced model, torque prediction improves for the actuated joints (HAA and HFE) but remains inaccurate at KFE due to the unmodeled four-bar closed-loop transmission.
    Nevertheless, the highlighted intervals show good agreement when the associated transmission mapping Jacobian is approximately unitary.
    (c) Identification with closed-loop kinematics enforces loop closure and resolves the KFE discrepancy, yielding consistent torque prediction across all joints.
}
\label{fig:comp_identified_model_torques}
\end{figure}

\Cref{fig:comp_identified_model_torques} quantifies the impact of modeling fidelity on torque prediction for a representative B1 leg trajectory.
With the nominal URDF parameters provided by Unitree (\Cref{fig:nominal_torques}), the model failed to explain the measured torques, with the largest discrepancy occurring at the KFE joint (approximately an order-of-magnitude mismatch).
After identifying inertial and friction parameters on a reduced model (\Cref{fig:identified_model_torques}), the torque prediction remained imperfect overall.
Specifically, noticeable mismatches persisted at the KFE joint, indicating that accurate modeling of the four-bar mechanism is important.
Note that, in the highlighted regions of~\Cref{fig:identified_model_torques} where the associated mapping Jacobian was approximately unitary, the predicted and measured torques aligned closely, whereas larger errors appeared when the Jacobian departed from this regime.
Therefore, enforcing the four-bar kinematics was important to close this gap as shown in~\Cref{fig:identified_model_torques_clk}.

\begin{figure}[t]\centering
    \includegraphics[width=0.67\linewidth]{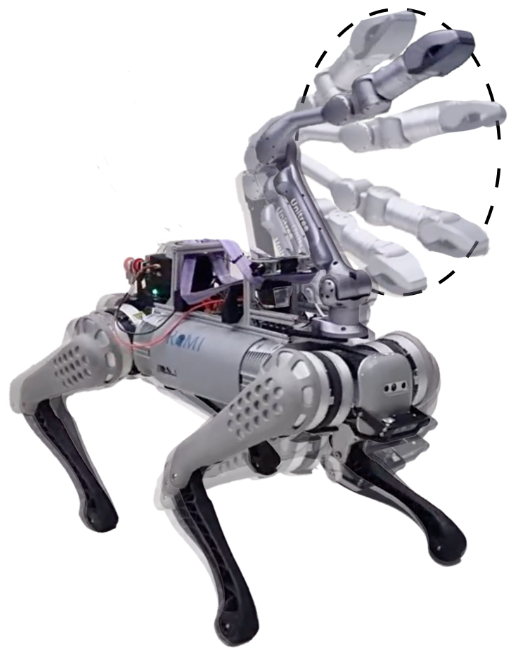}
    \includegraphics[width=.28\linewidth]{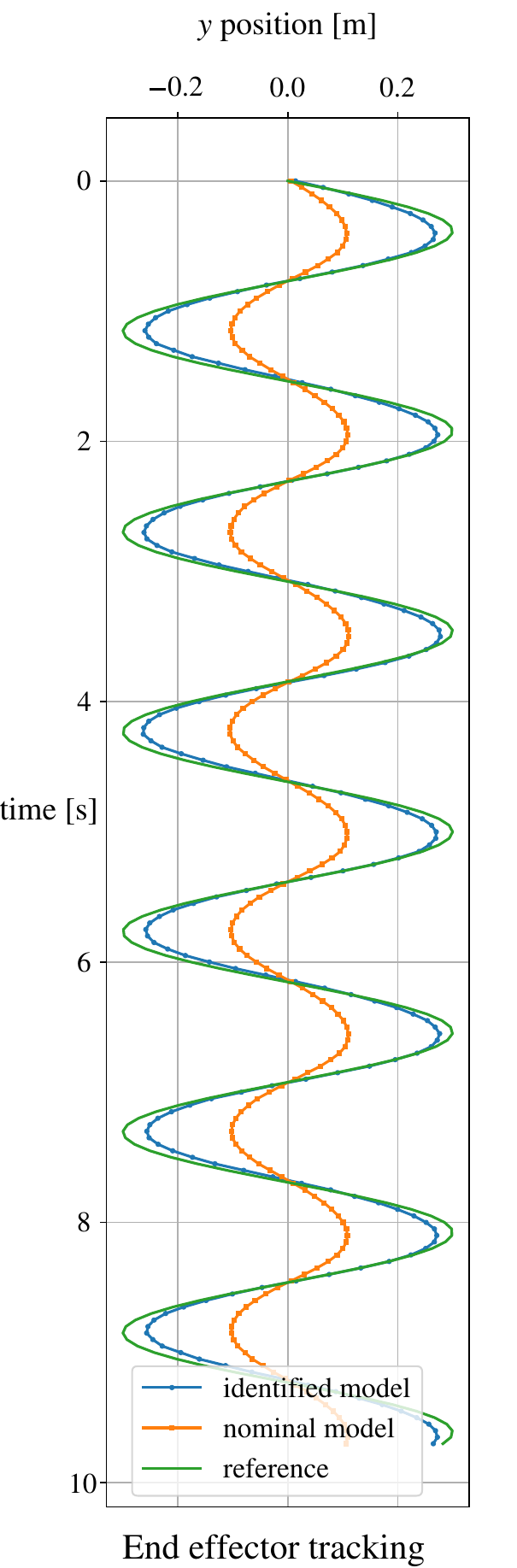}
    \caption{
        Top: Unitree B1 with a Z1 arm tracing a circular end–effector path while standing in four–point contact.  
        The arc highlights the commanded circle at the wrist; contact and loop–closure constraints are enforced during estimation, and the actuation/friction and energy terms are used to identify parameters without force/torque sensors.
        Bottom: End–effector tracking performance using the identified model (blue) versus the nominal URDF model (orange).
    }
    \label{fig:B1_Z1_circle_path}
\end{figure}

\begin{figure}[t]\centering
    \href{\video&t=319}{\includegraphics[width=1.0\linewidth]{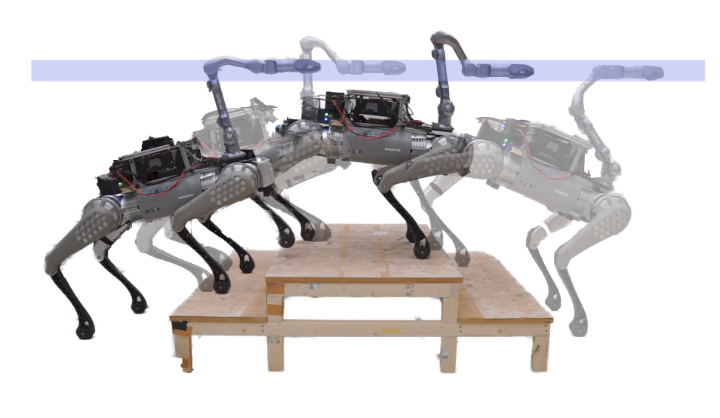}}
    \includegraphics[width=1.0\linewidth]{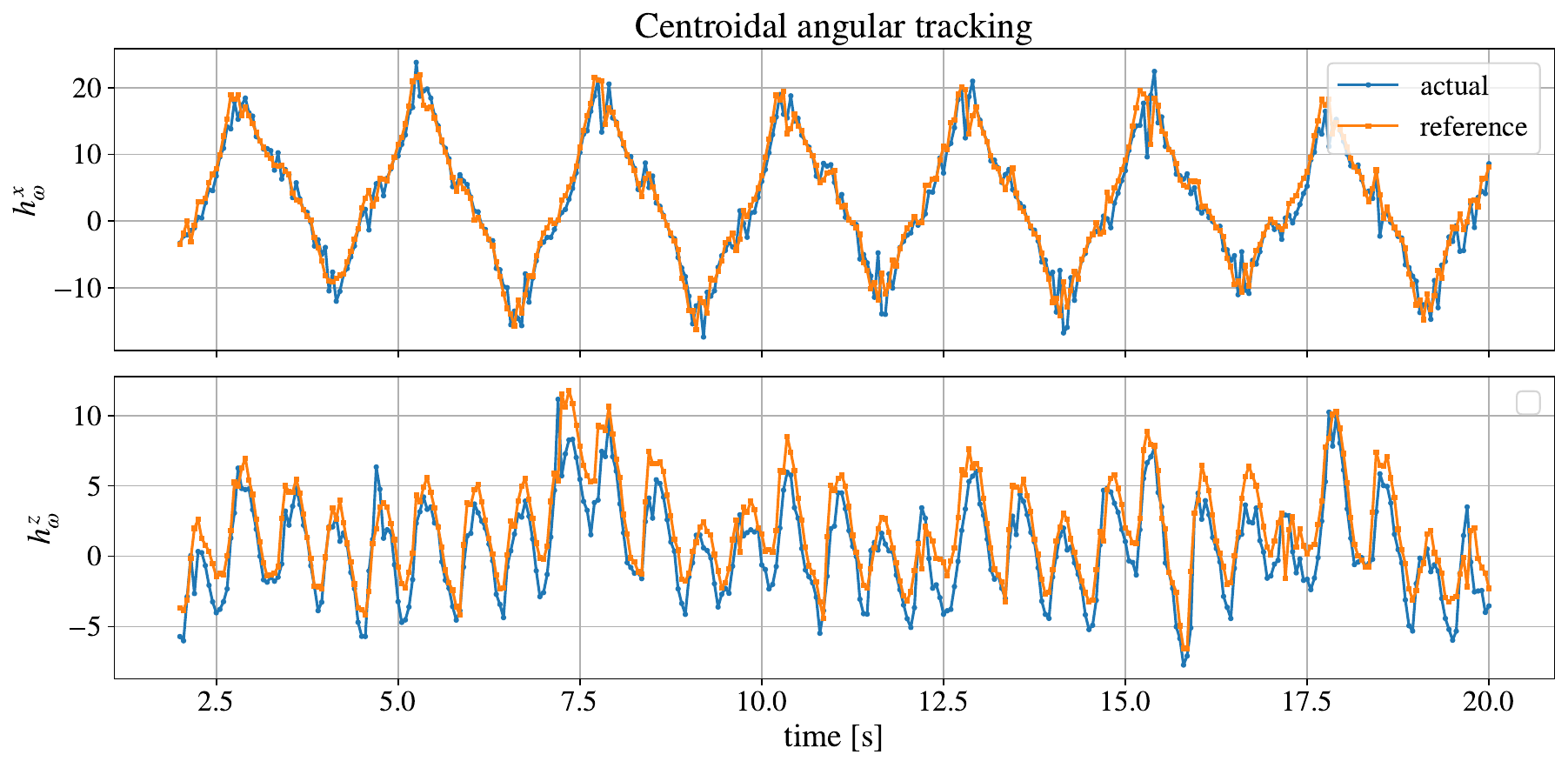}
    \caption{
        Top: Unitree B1 with a Z1 arm tracing a plane while climbing up the stairs. Bottom: tracking of the centroidal angular momentum.
    }
    \label{fig:B1_Z1_stairs}
\end{figure}

We then evaluated whether these improvements translated to task-level tracking.
\Cref{fig:B1_Z1_circle_path} shows a standing manipulation experiment in which the Z1 end-effector was commanded to trace a circular path while the robot remained in four-point contact.
Using the identified model improved end-effector tracking compared to the nominal URDF model.
Moreover, \Cref{fig:B1_Z1_stairs} illustrates a more dynamic scenario, in which the robot traced a plane while climbing stairs. In this case, the centroidal angular momentum tracking followed the reference closely, indicating that the identified model remained consistent under hybrid, contact-rich motion and improved whole-body tracking performance beyond the identification trajectory itself.
Other locomotion experiments are reported in~\Cref{fig:snapshots_results}

%% file: chapters/conclusions.tex
\section{Conclusion}\label{sec:conclusion}
We introduced a Bayesian approach for system and disturbance identification that embeds inverse dynamics as hard equality constraints and enforces state, contact, loop--closure, and parameter-consistency constraints.
Our Riccati solver preserves the~\gls{sysid} structure with a linear-time backward complexity, keeps numerical integration decoupled from the hybrid and closed-loop dynamics, and accommodates physically consistent inertial and friction parameterizations.
We also incorporated energy-based observations that couple measured actuation power to changes in mechanical energy, providing complementary information when actuation sensing is limited or indirect.

In simulation across a range of systems, we showed that inverse dynamics increases the quality of the identified parameters and disturbances compared to forward dynamics.
Specifically, it consistently achieved lower parameter error, with the largest gains in contact-rich and underactuated regimes.
Moreover, enforcing parameter constraints such as symmetry and total mass improved inertial estimates without sacrificing trajectory fit, while energy observations further enhanced identifiability---most notably reducing friction error---with minimal impact on trajectory accuracy.
On hardware, experiments trails on a Unitree B1 equipped with a Z1 arm showed the importance of simultaneously modeling foot contacts and closed-loop leg kinematics, recovering physically plausible inertial and friction parameters, and maintaining consistency with measured input-power balances during motion.


Overall, our work provides a principled and scalable route to physically consistent joint localization and system identification.
By unifying dynamics, constraints, and energy information within a single optimization program, it moves beyond decoupled pipelines and supports accurate model construction for agile manipulation and locomotion.

%% file: chapters/baumgarte_der.tex
\section{Derivatives of the Baumgarte stabilization.}
For either constraint type (bilateral contact or loop closure), the Baumgarte correction in constraint space is
\begin{align}\label{eq:Baumgarte}
\biasAcc^{*}(\pos,\vel) = -K_p\,\implicitConst{}(\pos)-K_d\,\dimplicitConst{}(\pos,\vel),
\end{align}
with Baumgarter gains $K_p,K_d\ge \zeroVec$ tuned by the user.

Using \Cref{eq:holonomic_constraints,eq:dholonomic_constraints}, the configuration derivatives of the configuration residuals are
\begin{align}
\frac{\partial \implicitContact}{\partial \pos}
&= \mathrm{Jlog6}\big(\!\spatialPlacement[1]{c}^{*}\ominus \spatialPlacement[1]{c}\big)\;\contactJac(\pos),\\
\frac{\partial \implicitLoop}{\partial \pos}ç
&= \mathrm{Jlog6}\big(\!\spatialPlacement[1]{2}\big)\;\loopJac(\pos),
\end{align}
where $\mathrm{Jlog6}$ denotes the Jacobian of the logarithm map from $\SE[3]$ to $\R^6$ (see e.g.,~\cite{sola2018micro}).
Since $\dimplicitContact=\contactJac(\pos)\vel$ and $\dimplicitLoop=\loopJac(\pos)\vel$, we have
$\frac{\partial \dimplicitConst{}}{\partial \pos} = \dmotionJac{}(\pos,\vel), \quad
\frac{\partial \dimplicitConst{}}{\partial \vel} = \motionJac{}(\pos)$,
where $\dmotionJac{}$ denotes the time derivative along $(\pos,\vel)$. The final expression becomes:
\begin{align}
\frac{\partial \biasAcc^{*}}{\partial \pos}
= -K^{}_p \frac{\partial \implicitConst{}}{\partial \pos} -K_d\,\dmotionJac{}(\pos,\vel),\quad
\frac{\partial \biasAcc^{*}}{\partial \vel}
= -K_d\,\motionJac{}(\pos).
\end{align}

%% file: chapters/excitation.tex
\section{Generation of optimal excitation trajectories}\label{sec:opt_traj}
Identifying the dynamic parameters requires to excite the system in a way that their are observable. 
This can be achieved by applying carefully designed input trajectories that maximize the sensitivity of the system's response to changes in the inertial and actuation parameters. 
Simillar to \cite{sturz2017parameter}, we formulate the following optimization problem to generate the optimal excitation trajectories:
\begin{align}\label{eq:opt_traj}
\min_{\{a_{i,\ell},\,b_{i,\ell}\}} \quad 
& \tfrac{1}{2}\frobnorm{Y_b(\pos,\vel,\acc)}^{2}+\tfrac{1}{2n_v}\,\frobnorm{Y_b(\pos,\vel,\acc)}^{2} &\\\nonumber
- w_v &\sum_{k=1}^{N-1}\| \vel[k]\|_2^2 &\\\nonumber
\text{s.t.}\quad \forall k=&1,\dots,N-1:\\\nonumber
\pos[0] = & \pos[N] = \pos[ini], \quad \vel[0] = \vel[N]= \mathbf{0}, \hspace{0.5em} \text{(initial conditions)} \\\nonumber
\acc[0] = &\acc[N]= \mathbf{0}, \\\nonumber
\pos[\min] \le & \pos[k] \le \pos[\max],\quad \vel[\min] \le \vel[k] \le \vel[\max], \hspace{0.5em} \text{(joint limits)} \\\nonumber
\Lambda_{ini} =& \text{FK}(\pos[k]), \hspace{9em} \text{(contact placemet)} \\\nonumber
\eff[\min] \le & \text{RNEA}(\pos[k],\vel[k],\acc[k], \boldsymbol{\lambda}_k) \le \eff[\max], \hspace{1em} \text{(torque limits)} \\\nonumber
\| \boldsymbol{\lambda}_k^T \| \leq & \mu \boldsymbol{\lambda}_k^N, \quad \boldsymbol{\lambda}_k^N \geq 0,  \hspace{7em} \text{(friction cone)}\\\nonumber
\text{with}\quad \forall i=&1,\dots,n_v,\ \forall k:\\\nonumber
\pos[k]^{i_{th}} =& \pos[ini]^{i_{th}} \oplus \sum_{\ell=1}^{L}\!\left(\frac{a_{i,\ell}}{\omega_\ell}\sin(\omega_\ell t_k)
                 - \frac{b_{i,\ell}}{\omega_\ell}\cos(\omega_\ell t_k)\right), \\\nonumber
\vel[k]^{i_{th}} =& \sum_{\ell=1}^{L}\!\left(a_{i,\ell}\cos(\omega_\ell t_k)
                 + b_{i,\ell}\sin(\omega_\ell t_k)\right), \\\nonumber
\acc[k]^{i_{th}} =& \sum_{\ell=1}^{L}\!\left(-a_{i,\ell}\omega_\ell\sin(\omega_\ell t_k)
                 + b_{i,\ell}\omega_\ell\cos(\omega_\ell t_k)\right), \\\nonumber
t_k =& \frac{k}{N-1}T,\quad k=0,\dots,N-1,
\end{align}
where $\pos[ini] \in \confManif$ is the initial configuration, $\Lambda_{ini} \in \mathbb{R}^{n_c\times 6}$ is the initial contact placement, \(Y_b=\big[\,Y(\pos[1],\vel[1],\acc[1]);\ldots;Y(\pos[N-1],\vel[N-1],\acc[N-1])\,\big]\in\mathbb{R}^{(N-1)n_v\times n_\theta}\) is the stacked of joint-torque regressors along the trajectory, and \(\|\cdot\|_F\) is the Frobenius norm. 

The term $\tfrac{1}{2}\frobnorm{Y_b(\pos,\vel,\acc)}^{2}+\tfrac{1}{2n_v}\,\frobnorm{Y_b(\pos,\vel,\acc)}^{2}$ is used to aproximate the condition number of the stacked regressor matrix, which is a measure of the sensitivity of the parameter estimates to noise in the data.
Contact forces are \(\boldsymbol{\lambda}_k=[\,\boldsymbol{\lambda}_k^{\hat{n}},\,\boldsymbol{\lambda}_k^{\hat{t}}\,]\) and satisfy the Coulomb cone \(\|\boldsymbol{\lambda}_k^{\hat{t}}\|_2\le \mu\,\boldsymbol{\lambda}_k^{\hat{n}}\) with \(\boldsymbol{\lambda}_k^{\hat{n}}\ge 0\). 
The constraint \(\Lambda_{\mathrm{ini}}=\mathrm{FK}(\pos[k])\) enforces a fixed contact pose (sticking). 
Joint and velocity limits are encoded by \(\pos[\min]\le \pos[k]\le \pos[\max]\) and \(\vel[\min]\le \vel[k]\le \vel[\max]\).
The time grid is \(t_k=\tfrac{k}{N-1}T\) for \(k=0,\ldots,N-1\); harmonic frequencies are \(\omega_\ell=\ell\,\omega_f\) for \(\ell=1,\ldots,L\); and \(w_v>0\) weights the speed-promoting term in the objective function.